\documentclass[11pt]{article}

\usepackage[final]{acl}
\usepackage{subcaption}
\usepackage{times}
\usepackage{latexsym}
\usepackage{multirow}

\usepackage{longtable}
\usepackage{pdflscape}  
\usepackage{array}
\usepackage{colortbl}

\usepackage{array}
\usepackage{colortbl}
\usepackage{xcolor}

\usepackage{multicol}
\usepackage{xcolor}

\usepackage{microtype}

\usepackage{inconsolata}

\usepackage{graphicx}

\usepackage{booktabs}
\usepackage{multirow}
\usepackage{graphicx}
\usepackage{xcolor}
\usepackage{colortbl}
\usepackage{amsmath}
\usepackage{amssymb}
\usepackage{geometry}
\usepackage{caption}
\usepackage{subcaption}
\usepackage{array}
\usepackage{makecell}
\usepackage{enumitem}

\usepackage{fontspec}
\usepackage{xcolor}
\usepackage{booktabs}
\usepackage{pifont}
\usepackage{array}
\usepackage{colortbl}
\usepackage{tabularx}
 
\newfontfamily\banglafont{LohitBengali}[
  Path = fonts/,
  Extension = .ttf,
  UprightFont = Lohit-Bengali,
]

\newcommand{\bn}[1]{{\banglafont #1}}
 
\definecolor{cA}{RGB}{31, 119, 180}
\definecolor{cB}{RGB}{44, 160, 44}
\definecolor{cC}{RGB}{255, 127, 14}
\definecolor{cD}{RGB}{214, 39, 40}
\definecolor{cdist}{RGB}{170, 40, 40}
\definecolor{growbg}{RGB}{253, 247, 247}
\definecolor{divcolor}{RGB}{190, 190, 190}
\definecolor{hdcolor}{RGB}{55, 55, 55}
 
\newcommand{\ma}[1]{\allowbreak\colorbox{cA!15}{\textcolor{cA}{\textbf{#1}}}\textsuperscript{\tiny\textcolor{cA}{\ding{192}}}\allowbreak}
\newcommand{\mb}[1]{\allowbreak\colorbox{cB!12}{\textcolor{cB!80!black}{\textbf{#1}}}\textsuperscript{\tiny\textcolor{cB!80!black}{\ding{193}}}\allowbreak}
\newcommand{\mc}[1]{\allowbreak\colorbox{cC!15}{\textcolor{cC!85!black}{\textbf{#1}}}\textsuperscript{\tiny\textcolor{cC!85!black}{\ding{194}}}\allowbreak}
\newcommand{\md}[1]{\allowbreak\colorbox{cD!12}{\textcolor{cD}{\textbf{#1}}}\allowbreak}
\newcommand{\dv}[1]{\textcolor{cdist}{\textbf{#1}}}

\usepackage{fontspec}
\usepackage{polyglossia}
\setmainlanguage{english}
\setotherlanguage{bengali}

\newfontfamily\bengalifont[
  Path = fonts/,
  Extension = .ttf,
  UprightFont = *-Regular,
  BoldFont = *-Bold,
  Script = Bengali,
  Renderer = HarfBuzz,
  Scale = 0.9
]{NotoSansBengali}

\usepackage[most]{tcolorbox}
\usepackage{listings}
\usepackage{xcolor}

\definecolor{promptbg}{RGB}{232, 240, 254}      
\definecolor{promptborder}{RGB}{127, 179, 255}    
\definecolor{jsonbg}{RGB}{80, 90, 112}
\definecolor{jsontext}{RGB}{171, 178, 191}

\newtcolorbox{promptbox}[1][]{
  enhanced,
  colback=promptbg,
  colframe=promptborder,
  fonttitle=\bfseries\sffamily\small,
  coltitle=black,
  title=#1,
  attach boxed title to top center={yshift=-2mm},
  boxed title style={
    colback=promptborder,
    colframe=promptborder,
    coltitle=white,
  },
  left=2mm,
  right=2mm,
  top=4mm,
  arc=2mm,
  boxrule=0.5pt,
  breakable,
  before skip=10pt,
  after skip=10pt,
}

\definecolor{verifybg}{RGB}{240, 253, 244}        
\definecolor{verifyborder}{RGB}{120, 163, 114}      
\definecolor{verifyjsonbg}{RGB}{80, 112, 89}       
\definecolor{verifyjsontext}{RGB}{203, 213, 225}  

\newtcolorbox{verifybox}[1][]{
  enhanced,
  colback=verifybg,
  colframe=verifyborder,
  fonttitle=\bfseries\sffamily\small,
  coltitle=white,
  title=#1,
  attach boxed title to top center={yshift=-2mm},
  boxed title style={
    colback=verifyborder,
    colframe=verifyborder,
    coltitle=white,
  },
  left=2mm,
  right=2mm,
  top=8mm,
  arc=2mm,
  boxrule=0.5pt,
  breakable,
  before skip=10pt,
  after skip=10pt,
}

\definecolor{dropred}{RGB}{220, 53, 69}
\definecolor{dropgreen}{RGB}{40, 167, 69}
\definecolor{headerblue}{RGB}{52, 73, 94}
\definecolor{reasoningmodel}{RGB}{230, 243, 255}

\definecolor{graphbg}{RGB}{240, 253, 250}        
\definecolor{graphborder}{RGB}{13, 148, 136}     
\definecolor{graphjsonbg}{RGB}{31, 61, 55}       
\definecolor{graphjsontext}{RGB}{209, 213, 219}  

\newtcolorbox{graphbox}[1][]{
  enhanced,
  colback=graphbg,
  colframe=graphborder,
  fonttitle=\bfseries\sffamily\small,
  coltitle=white,
  title=#1,
  attach boxed title to top center={yshift=-2mm},
  boxed title style={
    colback=graphborder,
    colframe=graphborder,
    coltitle=white,
  },
  left=2mm,
  right=2mm,
  top=8mm,
  arc=2mm,
  boxrule=0.5pt,
  breakable,
  before skip=10pt,
  after skip=10pt,
}

\definecolor{cotbg}{RGB}{240, 249, 255}         
\definecolor{cotborder}{RGB}{14, 165, 233}      
\definecolor{cotcontentbg}{RGB}{249, 250, 251}   
\definecolor{cotcontentborder}{RGB}{209, 213, 219} 

\newtcolorbox{cotbox}[1][]{
  enhanced,
  colback=cotbg,
  colframe=cotborder,
  fonttitle=\bfseries\sffamily\small,
  coltitle=white,
  title=#1,
  attach boxed title to top center={yshift=-2mm},
  boxed title style={
    colback=cotborder,
    colframe=cotborder,
    coltitle=white,
  },
  left=2mm,
  right=2mm,
  top=8mm,
  arc=2mm,
  boxrule=0.5pt,
  breakable,
  before skip=10pt,
  after skip=10pt,
}

\usepackage{xspace}
\newcommand{\ours}{$\dagger$\textsc{DAGGER}\xspace} 
\newcommand{\ourdb}{\textsc{DistractMath-BN}\xspace} 
\newcommand{\DataReleaseURL}[0]{\url{https://dipta007.github.io/DAGGER/}}

\usepackage{xcolor}
\usepackage{amssymb}
\definecolor{ourgold}{HTML}{DAA520}

\title{\ours: Distractor-Aware Graph Generation for Executable Reasoning in Math Problems}

\author{
    Zabir Al Nazi\hspace{.1em}$^{\,1}$,\quad
    Shubhashis Roy Dipta\hspace{.1em}$^{\,2}$,\quad
    Sudipta Kar\hspace{.1em}$^{\,3}$ \vspace{1mm}\\
    $^1$\textbf{University of California, Riverside} \\
    $^2$\textbf{University of Maryland, Baltimore County}\\
    $^3$\textbf{Oracle Health AI}\vspace{1mm}\\
    \texttt{znazi002@ucr.edu}, \texttt{sroydip1@umbc.edu}, \texttt{sudipta.k.kar@oracle.com}
}

\begin{document}
\maketitle

\begin{abstract}
Chain-of-Thought (CoT) prompting is widely adopted for
mathematical problem solving, including in low-resource languages,
yet its robustness under irrelevant context - a known vulnerability
in English - remains unexamined in such settings. To systematically probe this challenge, we introduce \textbf{\ourdb}, a Bangla benchmark that augments MGSM and MSVAMP with semantically coherent but computationally irrelevant information. Evaluating seven models ranging from 3B to 12B parameters, we observe substantial performance degradation under distractors: standard models drop up to 41 points, while reasoning-specialized models decline 14--20 points despite consuming 5× more tokens. We propose \ours, which reformulates mathematical problem solving as executable computational graph generation with dedicated modeling of distractor nodes. Fine-tuning Gemma-3 models using Supervised Fine-Tuning followed by Group Relative Policy Optimization achieves comparable aggregate accuracy across original and distractor-augmented splits while using 89\% fewer tokens than reasoning models. Importantly, this robustness emerges without explicit training on distractor-augmented examples. Our results suggest that enforcing structured intermediate representations can improve robustness and inference efficiency in mathematical understanding compared to free-form reasoning, particularly in noisy, low-resource settings.\footnote{\DataReleaseURL}
\end{abstract}

\begin{figure}[!t]
    \centering
    \includegraphics[width=0.94\columnwidth]{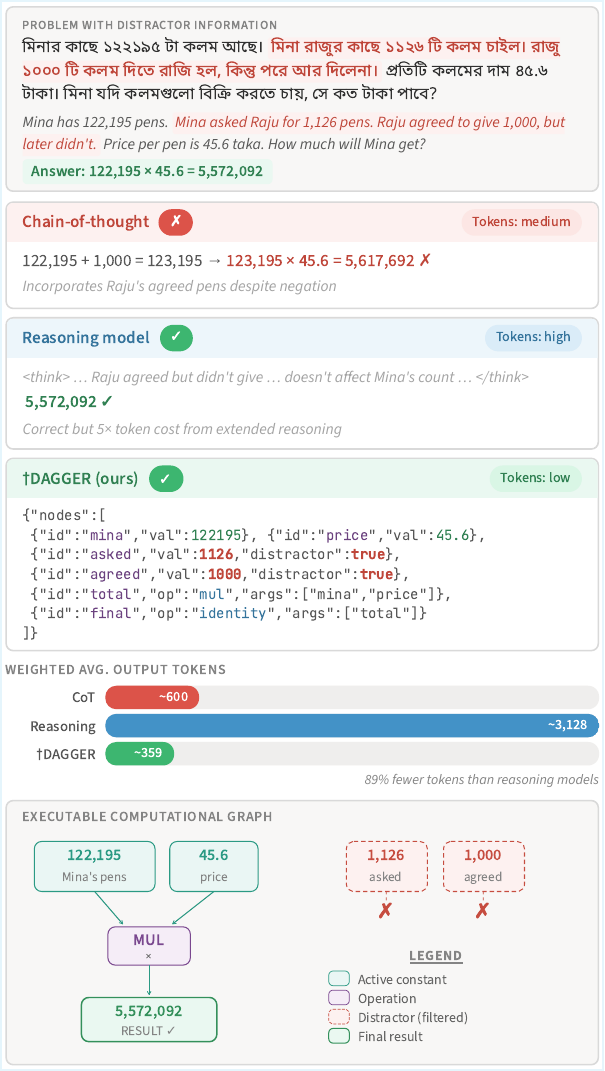}
    \caption{Comparing CoT, reasoning, and \ours approaches on a distractor-augmented Bangla MWP.}
    
    \label{fig:single_column}
\end{figure}
\section{Introduction}

Chain-of-Thought (CoT) prompting has become a standard approach for mathematical word problem (MWP) solving in large language models (LLMs), enabling models to generate intermediate reasoning steps prior to producing final answers~\cite{wei2022chain, cheng2025revisiting}. This paradigm has been widely adopted across languages and datasets, including recent work demonstrating that CoT-based prompting provides strong baselines for Bangla mathematical reasoning benchmarks ~\cite{shi2022language, ahmed2025bennumeval}. Reasoning-specialized models further extend this paradigm by allocating additional computation to multi-step reasoning through longer or iterative generations \cite{yang2025qwen3, luong2025towards} but at substantially higher token budgets.



Despite these advances, mathematical problem solving outside curated benchmarks frequently involves semantically coherent but computationally irrelevant information - \emph{distractors} - such as quantities about related entities, attributes in orthogonal dimensions, or events that are negated or hypothetical (Fig.~\ref{fig:single_column}). Prior work has established this vulnerability in English: \citet{shi2023large} showed irrelevant sentences degrade reasoning significantly, and subsequent studies confirm this fragility under adversarial settings~\citep{li2023you, yang2025iterative}. However, distractor robustness in low-resource languages remains under-examined, and the \emph{computational cost} of achieving robustness is largely overlooked. Reasoning-specialized models trained with reinforcement learning~\citep{guo2025deepseek} show improved robustness but require 10-100$\times$ more tokens at inference~\citep{snell2025scaling}, raising a practical question: \textit{\textbf{can we achieve distractor robustness without the computational overhead of extended reasoning, particularly in resource-constrained settings where both accuracy and efficiency matter?}}

\paragraph{Research Questions}

Low-resource languages like Bangla show greater vulnerability to irrelevant context during math reasoning \cite{sobhani2025mathmist, prama2025banglamath}. While CoT prompting improves English math reasoning \cite{wei2022chain}, it exhibits brittleness to distractors \cite{shi2023large}. Reasoning models trained using Group Relative Policy Optimization (GRPO) achieve substantial gains in mathematical problem-solving \cite{guo2025deepseek} but require more tokens \cite{snell2025scaling}, raising deployment concerns. 

Recent works suggest reasoning traces produced by reasoning models can be structured as \textit{\textbf{graphs}} \cite{minegishi2025topology}, potentially enabling more efficient reasoning \cite{yu2024flow}. These observations motivate us to investigate the following research questions.
\textbf{RQ1}: whether CoT-based reasoning remains robust on distractor-augmented Bangla MWPs; \textbf{RQ2}: whether structured reasoning frameworks can be distractor-aware \textit{\textbf{without targeted}} distractor-type training; and \textbf{RQ3}: the computational trade-offs between robustness and token efficiency.

\paragraph{Contributions}
We make the following contributions in this paper.
\vspace{-5pt}
\begin{itemize}[leftmargin=12pt]
    \item We introduce \ourdb, a novel Bangla benchmark of 3,685 distractor-augmented problems derived from MGSM and MSVAMP validated by GPT and human expert. The benchmark systematically incorporates three classes of interference - Related Entity Distractors, Orthogonal Attribute Distractors, and Null-Effect Event Distractors \textit{(details in Section~\ref{sec:dataset}, examples in Appendix~\ref{app:examples})}.
    \vspace{-5pt}
    \item Thorough evaluation of seven open-weight models ranging from 3B to 12B parameters, we show that CoT-based reasoning degrades substantially in the presence of distractors.
    \vspace{-5pt}
    \item Our proposed framework \ours reformulates mathematical problem solving as executable computational graph generation with explicit modeling of distractor nodes.
    \vspace{-5pt}
    \item We show substantial accuracy and token efficiency over long-form CoT by fine-tuning Gemma-3 (SFT+GRPO) using \ours framework for improved graph generation. 
\end{itemize}
\vspace{-5pt}
Overall, our findings indicate that robustness to distractor information is not guaranteed by longer reasoning traces alone. Instead, enforcing structured intermediate representations provides a practical mechanism for improving both robustness and inference efficiency in mathematical reasoning, particularly in noisy, low-resource settings.

\section{Related Work}

\paragraph{Chain-of-Thought Reasoning}
CoT prompting~\cite{wei2022chain} enables large language models (LLMs) to generate intermediate reasoning steps, substantially improving performance on mathematical word problems. Subsequent variants such as self-consistency~\cite{wang2023selfconsistency}, and least-to-most prompting~\cite{zhou2023least} further enhance reasoning accuracy. In multilingual and low-resource settings, recent studies report that CoT-based prompting remains a strong baseline for Bangla mathematical reasoning benchmarks \cite{ahmed2025bennumeval}. 

\paragraph{Reasoning Robustness}
Prior work on mathematical problem solving has shown that, while CoT prompting can improve accuracy on math word problems, the resulting step-by-step rationales are often brittle and not reliably indicative of the model’s underlying computation - models may produce convincing but unfaithful explanations and still fail at exact or algorithmic calculation \cite{turpin2023language, sprague2024cot}.
More closely related are studies showing that irrelevant or misleading context can significantly degrade model performance in question answering and reasoning tasks~\cite{shi2023large, roydipta2025depresuppose, nazi2026omdbench}. 

\paragraph{Structured Reasoning}
  Program-Aided Language Models (PAL)~\cite{gao2023pal} and Program-of-Thoughts (PoT)~\cite{chen2022program} translate reasoning into executable programs, offloading computation to external interpreters such as Python. While effective for numerical accuracy, these methods generate code sequentially and treat all input tokens uniformly, offering limited mechanisms to isolate computationally irrelevant context. Related work on graph-based reasoning and planning~\cite{yao2023think, besta2024graph} focuses primarily on knowledge graphs or abstract decision processes rather than executable computational structures for math. In particular, Graph-of-Thoughts~\citep{besta2024graph} 
models reasoning as a graph over alternative 
\emph{thought strategies} explored via multiple LLM 
calls ($\sim$6 per problem), whereas \dag\textsc{Dagger} 
generates a single executable \emph{computational} graph 
in one call, with explicit distractor annotations 
absent from all prior approaches. Particularly, \ours generates executable computational graphs with dedicated modeling of distractor nodes, enabling structural separation between relevant computation and irrelevant information prior to execution.

\begin{table}[t]
\centering
\scriptsize
\renewcommand{\arraystretch}{1.15}
\setlength{\tabcolsep}{3pt}

\begin{tabularx}{\columnwidth}{
  !{\color{divcolor}\vrule width 0.4pt}
  p{0.7cm}
  !{\color{divcolor}\vrule width 0.4pt}
  X
  !{\color{divcolor}\vrule width 0.4pt}
}
 
\arrayrulecolor{hdcolor}
\toprule
\rowcolor{gray!8}
\textbf{Type} & \textbf{Problem Text} \\
\midrule
 
\textbf{Orig.}
\newline {\scriptsize\textbf{Ans: 8}}
&
\bn{\ma{জুলিয়া} \mc{সোমবার} \mb{১৮} জন \mb{ছেলেমেয়ের} সঙ্গে ট্যাগ খেলেন। \mc{মঙ্গলবার} তিনি \mb{১০} জন ছেলেমেয়ের সঙ্গে ট্যাগ খেলেন। সোমবার তিনি কতজন ছেলেমেয়ের সঙ্গে আরও খেলেন?}
\par\vspace{1pt}
{\tiny\color{gray!60}\textit{%
\ma{Julia} played tag with \mb{18} \mb{children} on \mc{Mon}. On \mc{Tue} she played with \mb{10} children. How many more on Mon?}}
\\
 
\arrayrulecolor{cD!50}\specialrule{1pt}{0pt}{0pt}\arrayrulecolor{hdcolor}
 
\cellcolor{growbg}
{\scriptsize\textbf{\ding{192} RED}}
\newline {\tiny Related}\newline{\tiny Entity}
&
\cellcolor{growbg}
\textcolor{cdist}{\bn{%
\mc{মঙ্গলবার} মাঠে আরও \dv{১৫} জন \mb{ছেলেমেয়ে} খেলতে এসেছিল। তার ছোট \dv{বোন} বুধবার \dv{১২} জন \mb{ছেলেমেয়ের} সঙ্গে লুকোচুরি খেলেছিল।}}
\par\vspace{1pt}
{\tiny\color{gray!60}\textit{%
\textcolor{cdist}{\dv{15} more \mb{children} came on \mc{Tue}. Her \dv{sister} played with \dv{12} \mb{children} on Wed.}}}
\par\vspace{0.5pt}
{\tiny\color{gray!50}\textit{%
$\hookrightarrow$ \textsuperscript{\tiny\ding{193}} same item (children), \textsuperscript{\tiny\ding{194}} same time (Tue); different actors}}
\\[-1pt]
 
\arrayrulecolor{divcolor}\specialrule{0.4pt}{0pt}{0pt}\arrayrulecolor{hdcolor}
 
\cellcolor{growbg}
{\scriptsize\textbf{\ding{193} OAD}}
\newline {\tiny Ortho.}\newline{\tiny Attrib.}
&
\cellcolor{growbg}
\textcolor{cdist}{\bn{%
\mc{সোমবার} খেলতে \dv{১ ঘণ্টা} সময় লেগেছিল। \mc{মঙ্গলবার} খেলতে \dv{৪৫ মিনিট} সময় লেগেছিল।}}
\par\vspace{1pt}
{\tiny\color{gray!60}\textit{%
\textcolor{cdist}{It took \dv{1 hour} on \mc{Mon}. It took \dv{45 min} on \mc{Tue}.}}}
\par\vspace{0.5pt}
{\tiny\color{gray!50}\textit{%
$\hookrightarrow$ \textsuperscript{\tiny\ding{194}} same events (Mon/Tue); orthogonal dim.\ (time $\perp$ count)}}
\\[-1pt]
 
\arrayrulecolor{divcolor}\specialrule{0.4pt}{0pt}{0pt}\arrayrulecolor{hdcolor}
 
\cellcolor{growbg}
{\scriptsize\textbf{\ding{194} NEED}}
\newline {\tiny Null-Eff.}\newline{\tiny Event}
&
\cellcolor{growbg}
\textcolor{cdist}{\bn{%
\ma{জুলিয়া} ভেবেছিলেন আরও \dv{৫} জন ছেলেমেয়েকে \mc{সোমবার} খেলায় ডাকবেন, \md{কিন্তু পরে আর ডাকলেন না}। যদি সোমবার বৃষ্টি না হতো, আরও \dv{৩} জন যোগ দিত, \md{কিন্তু শেষ পর্যন্ত আসেনি}।}}
\par\vspace{1pt}
{\tiny\color{gray!60}\textit{%
\textcolor{cdist}{\ma{Julia} thought of inviting \dv{5} more on \mc{Mon}, \md{but didn't}. If no rain on Mon, \dv{3} would've joined, \md{but didn't come}.}}}
\par\vspace{0.5pt}
{\tiny\color{gray!50}\textit{%
$\hookrightarrow$ \textsuperscript{\tiny\ding{192}} same subject (Julia), \textsuperscript{\tiny\ding{194}} same time (Mon); \md{negation} cancels effects}}
\\[-1pt]
 
\bottomrule
\end{tabularx}
 
\vspace{2pt}
\centering
{\tiny
\ma{subj.}\;\mb{entity}\;\mc{event}\;\md{negation}\;
\textcolor{cdist}{red}\,=\,distractor\;
\textcolor{cdist}{\textbf{bold}}\,=\,new val.
}
\caption{Distractor augmentation (\ourdb). Markers \textsuperscript{\tiny\textcolor{cA}{\ding{192}}}\,subject, \textsuperscript{\tiny\textcolor{cB!80!black}{\ding{193}}}\,entity, \textsuperscript{\tiny\textcolor{cC!85!black}{\ding{194}}}\,event link distractors to original anchors. Answer ($18{-}10{=}8$) is unchanged.}
\label{tab:distractor-anatomy}
 
\end{table}

\paragraph{Bangla Mathematical Reasoning}
MGSM~\cite{shi2022language} and MSVAMP~\cite{patel2021nlp} provide multilingual benchmarks for mathematical reasoning, and recent Bangla work builds instruction-tuned models \cite{zehady2026banglallama} and a curriculum-GRPO math reasoner \cite{dipta2026ganitllmdifficultyawarebengalimathematical} evaluated on their Bangla splits. We extend these benchmarks with systematic distractor augmentation enabling controlled evaluation of reasoning robustness under semantically plausible but computationally irrelevant context in low-resource language \textit{(Bangla)}.

\section{\ourdb Dataset} \label{sec:dataset}


To systematically evaluate the robustness of mathematical reasoning under realistic noise conditions, we construct \ourdb, a distractor-augmented benchmark derived from Bangla subset of MGSM \textit{(250 problems)} and MSVAMP \textit{(1000 problems)}. For each problem in the source datasets, we generate \textit{three} distinct augmentation variants, each targeting a different cognitive interference mechanism.

\paragraph{Related Entity Distractors (RED)} augments problems with numerical information about the same object type as the query target, but associated with different entities or contexts. These sentences introduce quantities pertaining to related individuals, locations, or situations that are irrelevant to solving the problem. The distractors test whether models can distinguish between numerically relevant facts about the focal entity versus similar facts about peripheral entities (details in \ref{app:red_prompt}). 

\paragraph{Orthogonal Attribute Distractors (OAD)} adds supplementary properties to objects already present in the problem, where these properties are measured in dimensions orthogonal to the queried attribute. This tests whether models can identify which numerical attributes contribute to the solution when multiple properties of identical entities are specified (details in \ref{app:oad_prompt}).

\paragraph{Null-Effect Event Distractors (NEED)} introduces descriptions of events involving the problem's main entities that have zero net impact on the final answer. These include actions that were planned but not executed, operations that were performed then reversed, or counterfactual scenarios explicitly marked as unrealized. NEED tests whether models correctly interpret linguistic markers of negation, cancellation, and hypotheticality when processing event sequences (details in \ref{app:need_prompt}).

\paragraph{Design Principles and Ecological Validity} Our distractors are 
\emph{structurally anchored} to the original problem unlike prior work that evaluates robustness using 
randomly inserted or semantically unrelated 
sentences~\cite{shi2023large}. 
As illustrated in Table~\ref{tab:distractor-anatomy}, 
each distractor sentence is grounded in existing 
entities, subjects, or events from the problem. 
This closely aligns with how real-world math problems naturally 
contain plausible but computationally irrelevant 
information. This ensures \textit{\textbf{naturalness of the augmented problems}}: 
because distractors reference existing entities, 
events, and attributes rather than introducing 
unrelated context, they are semantically integrated 
into the problem and cannot be filtered through 
surface-level heuristics, unlike randomly appended 
irrelevant sentences. 
To preserve problem integrity, our generation pipeline 
uses structured JSON prompting 
(Apps.~\ref{app:prompts}) that instructs the model 
to produce \emph{only candidate distractor sentences 
with specified insertion positions}, rather than 
paraphrasing or rewriting the original problem. 
This ensures the core problem text and its solution 
remain untouched while the augmented problem reads 
as a natural, coherent Bangla math problem. 
Full examples across both MGSM and MSVAMP for all 
three distractor types are provided in 
Table~\ref{app:examples}.

\paragraph{Quality Assurance} Each augmented problem undergoes a two-stage verification pipeline: (1) automated validation using \texttt{GPT-4.1} to confirm answer preservation (prompt in \ref{app:gpt_verify}), and (2) human expert review by a native Bangla speaker with mathematical training. Only samples passing both stages are retained, yielding 738 verified augmented problems for MGSM and 2947 for MSVAMP.

We report the distribution of distractor nodes across 
training and evaluation data in 
Table~\ref{app:distractor_dist}: augmented evaluation 
problems contain 2.7 distractors on average, while 
training data averages only 0.21 - these are naturally 
occurring distractors identified by the teacher model 
(\texttt{GPT-4.1}) during graph generation, not artificially 
injected. No distractor augmentation is applied to 
training data; this gap is intentional, as it allows 
us to test whether learning to model distractor nodes 
through the graph structure alone, refined via GRPO, 
is sufficient to induce robustness under substantially 
heavier distractor loads than seen during training.

\section{Methodology}
\subsection{Computational Graph Framework}

\begin{table*}[t]
\centering
\tiny
\renewcommand{\arraystretch}{1.15}
\setlength{\tabcolsep}{2pt}
\begin{tabular}{@{}l *{5}{cc} *{2}{cc} *{6}{cc} @{}}
\toprule
& \multicolumn{10}{c}{\cellcolor{gray!15}\textbf{LLM (5-shot CoT)}} & \multicolumn{4}{c}{\cellcolor{blue!15}\textbf{Reasoning Models}} & \multicolumn{12}{c}{\cellcolor{green!15}\textbf{\ours (Gemma 3)}} \\
\cmidrule(lr){2-11} \cmidrule(lr){12-15} \cmidrule(lr){16-27}
& \multicolumn{4}{c}{Qwen 2.5} & \multicolumn{2}{c}{LLaMA 3} & \multicolumn{4}{c}{Gemma 3} 
& \multicolumn{4}{c}{Qwen 3} 
& \multicolumn{4}{c}{Gemma 3 w/o train.} & \multicolumn{4}{c}{Gemma 3 SFT} & \multicolumn{4}{c}{Gemma 3 GRPO} \\[-3pt]
\cmidrule(lr){2-5} \cmidrule(lr){6-7} \cmidrule(lr){8-11} 
\cmidrule(lr){12-15}
\cmidrule(lr){16-19} \cmidrule(lr){20-23} \cmidrule(lr){24-27}
& \multicolumn{2}{c}{3B} & \multicolumn{2}{c}{7B} & \multicolumn{2}{c}{8B} & \multicolumn{2}{c}{4B} & \multicolumn{2}{c}{12B} 
& \multicolumn{2}{c}{4B} & \multicolumn{2}{c}{8B} 
& \multicolumn{2}{c}{4B} & \multicolumn{2}{c}{12B} & \multicolumn{2}{c}{4B} & \multicolumn{2}{c}{12B} & \multicolumn{2}{c}{4B} & \multicolumn{2}{c}{12B} \\[-3pt]
\cmidrule(lr){2-3} \cmidrule(lr){4-5} \cmidrule(lr){6-7} \cmidrule(lr){8-9} \cmidrule(lr){10-11} 
\cmidrule(lr){12-13} \cmidrule(lr){14-15}
\cmidrule(lr){16-17} \cmidrule(lr){18-19} \cmidrule(lr){20-21} \cmidrule(lr){22-23} \cmidrule(lr){24-25} \cmidrule(lr){26-27}
\textbf{Metric} & MG & MS & MG & MS & MG & MS & MG & MS & MG & MS 
& MG & MS & MG & MS 
& MG & MS & MG & MS & MG & MS & MG & MS & MG & MS & MG & MS \\
\midrule
Accuracy (Original) 
& \cellcolor{gray!8}38.4 & \cellcolor{gray!8}53.7 & \cellcolor{gray!8}26.4 & \cellcolor{gray!8}53.9 & \cellcolor{gray!8}49.2 & \cellcolor{gray!8}54.7 & \cellcolor{gray!8}54.8 & \cellcolor{gray!8}60.7 & \cellcolor{gray!8}76.8 & \cellcolor{gray!8}72.3 
& \cellcolor{blue!8}\textbf{82.8} & \cellcolor{blue!8}78.2 & \cellcolor{blue!8}\textbf{88.0} & \cellcolor{blue!8}\textbf{81.1} 
& \cellcolor{green!8}24.8 & \cellcolor{green!8}57.2 & \cellcolor{green!8}60.4 & \cellcolor{green!8}74.2 & \cellcolor{green!8}40.4 & \cellcolor{green!8}65.0 & \cellcolor{green!8}70.0 & \cellcolor{green!8}76.8 & \cellcolor{green!8}54.8 & \cellcolor{green!8}70.3 & \cellcolor{green!8}78.4 & \cellcolor{green!8}\textbf{78.8} \\
Accuracy (+Distractor) 
& \cellcolor{gray!8}13.7 & \cellcolor{gray!8}13.0 & \cellcolor{gray!8}23.2 & \cellcolor{gray!8}35.8 & \cellcolor{gray!8}22.2 & \cellcolor{gray!8}26.9 & \cellcolor{gray!8}26.3 & \cellcolor{gray!8}28.8 & \cellcolor{gray!8}54.3 & \cellcolor{gray!8}48.7 
& \cellcolor{blue!8}62.9 & \cellcolor{blue!8}63.1 & \cellcolor{blue!8}\textbf{70.5} & \cellcolor{blue!8}\textbf{66.9} 
& \cellcolor{green!8}9.9 & \cellcolor{green!8}31.7 & \cellcolor{green!8}47.7 & \cellcolor{green!8}61.1 & \cellcolor{green!8}25.1 & \cellcolor{green!8}42.4 & \cellcolor{green!8}56.8 & \cellcolor{green!8}\textbf{65.4} & \cellcolor{green!8}31.4 & \cellcolor{green!8}42.9 & \cellcolor{green!8}64.0 & \cellcolor{green!8}\textbf{66.8} \\
$\Delta$ Accuracy ($\downarrow$) 
& \cellcolor{gray!8}\color{red}24.7 & \cellcolor{gray!8}\color{red}40.7 & \cellcolor{gray!8}\color{red}\textbf{3.2} & \cellcolor{gray!8}\color{red}18.1 & \cellcolor{gray!8}\color{red}27.0 & \cellcolor{gray!8}\color{red}27.8 & \cellcolor{gray!8}\color{red}28.5 & \cellcolor{gray!8}\color{red}31.9 & \cellcolor{gray!8}\color{red}22.5 & \cellcolor{gray!8}\color{red}23.6 
& \cellcolor{blue!8}\color{red}19.9 & \cellcolor{blue!8}\color{red}15.1 & \cellcolor{blue!8}\color{red}17.5 & \cellcolor{blue!8}\color{red}14.2 
& \cellcolor{green!8}\color{red}14.9 & \cellcolor{green!8}\color{red}25.5 & \cellcolor{green!8}\color{red}\textbf{12.7} & \cellcolor{green!8}\color{red}13.1 & \cellcolor{green!8}\color{red}15.3 & \cellcolor{green!8}\color{red}22.7 & \cellcolor{green!8}\color{red}13.2 & \cellcolor{green!8}\color{red}\textbf{11.4} & \cellcolor{green!8}\color{red}23.4 & \cellcolor{green!8}\color{red}27.4 & \cellcolor{green!8}\color{red}14.4 & \cellcolor{green!8}\color{red}\textbf{12.0} \\
\midrule
\rowcolor{yellow!10}
\textbf{Weighted Avg. Acc.} & \multicolumn{2}{c}{\cellcolor{gray!8}22.6} & \multicolumn{2}{c}{\cellcolor{gray!8}37.1} & \multicolumn{2}{c}{\cellcolor{gray!8}33.0} & \multicolumn{2}{c}{\cellcolor{gray!8}36.2} & \multicolumn{2}{c}{\cellcolor{gray!8}55.7} 
& \multicolumn{2}{c}{\cellcolor{blue!8}\textbf{67.1}} & \multicolumn{2}{c}{\cellcolor{blue!8}\textbf{71.4}} 
& \multicolumn{2}{c}{\cellcolor{green!8}33.3} & \multicolumn{2}{c}{\cellcolor{green!8}61.7} & \multicolumn{2}{c}{\cellcolor{green!8}44.3} & \multicolumn{2}{c}{\cellcolor{green!8}\textbf{66.7}} & \multicolumn{2}{c}{\cellcolor{green!8}47.3} & \multicolumn{2}{c}{\cellcolor{green!8}\textbf{69.4}} \\
\midrule
Tokens (Original) 
& \cellcolor{gray!8}536 & \cellcolor{gray!8}336 & \cellcolor{gray!8}560 & \cellcolor{gray!8}413 & \cellcolor{gray!8}295 & \cellcolor{gray!8}349 & \cellcolor{gray!8}706 & \cellcolor{gray!8}587 & \cellcolor{gray!8}582 & \cellcolor{gray!8}518 
& \cellcolor{blue!8}2068 & \cellcolor{blue!8}1767 & \cellcolor{blue!8}2074 & \cellcolor{blue!8}1802 
& \cellcolor{green!8}620 & \cellcolor{green!8}\textbf{235} & \cellcolor{green!8}376 & \cellcolor{green!8}\textbf{225} & \cellcolor{green!8}392 & \cellcolor{green!8}\textbf{231} & \cellcolor{green!8}407 & \cellcolor{green!8}\textbf{222} & \cellcolor{green!8}371 & \cellcolor{green!8}252 & \cellcolor{green!8}411 & \cellcolor{green!8}237 \\
Tokens (+Distractor) ($\uparrow$) 
& \cellcolor{gray!8}784 & \cellcolor{gray!8}710 & \cellcolor{gray!8}707 & \cellcolor{gray!8}662 & \cellcolor{gray!8}413 & \cellcolor{gray!8}399 & \cellcolor{gray!8}680 & \cellcolor{gray!8}743 & \cellcolor{gray!8}659 & \cellcolor{gray!8}612 
& \cellcolor{blue!8}3842 & \cellcolor{blue!8}3522 & \cellcolor{blue!8}3662 & \cellcolor{blue!8}3534 
& \cellcolor{green!8}911 & \cellcolor{green!8}\textbf{308} & \cellcolor{green!8}484 & \cellcolor{green!8}\textbf{299} & \cellcolor{green!8}538 & \cellcolor{green!8}373 & \cellcolor{green!8}500 & \cellcolor{green!8}\textbf{324} & \cellcolor{green!8}458 & \cellcolor{green!8}\textbf{353} & \cellcolor{green!8}519 & \cellcolor{green!8}356 \\
\midrule
\rowcolor{yellow!10}
\textbf{Weighted Out. Tokens} & \multicolumn{2}{c}{\cellcolor{gray!8}636} & \multicolumn{2}{c}{\cellcolor{gray!8}613} & \multicolumn{2}{c}{\cellcolor{gray!8}386} & \multicolumn{2}{c}{\cellcolor{gray!8}700} & \multicolumn{2}{c}{\cellcolor{gray!8}599} 
& \multicolumn{2}{c}{\cellcolor{blue!8}3140} & \multicolumn{2}{c}{\cellcolor{blue!8}3128} 
& \multicolumn{2}{c}{\cellcolor{green!8}399} & \multicolumn{2}{c}{\cellcolor{green!8}\textbf{316}} & \multicolumn{2}{c}{\cellcolor{green!8}370} & \multicolumn{2}{c}{\cellcolor{green!8}\textbf{334}} & \multicolumn{2}{c}{\cellcolor{green!8}\textbf{349}} & \multicolumn{2}{c}{\cellcolor{green!8}\textbf{359}} \\
\bottomrule
\end{tabular}
\vspace{0.5mm}

{\tiny MG = MGSM, MS = MSVAMP; w/o train. = without training; GRPO = SFT$\rightarrow$GRPO; $\downarrow$ = lower is better, $\uparrow$ = higher indicates longer generation. Top 4 results per row are shown in bold.}

\vspace{1mm}
\caption{\textbf{Impact of Distractor Augmentation on Mathematical Reasoning across MGSM and MSVAMP (Bangla).} Color coding distinguishes 3 model categories. Weighted metrics provide overall performance.}
\label{tab:merged_results}
\vspace{1mm}
\end{table*}

\begin{figure*}[t]
 \centering
 \begin{subfigure}[t]{0.32\textwidth}
   \includegraphics[width=\linewidth]{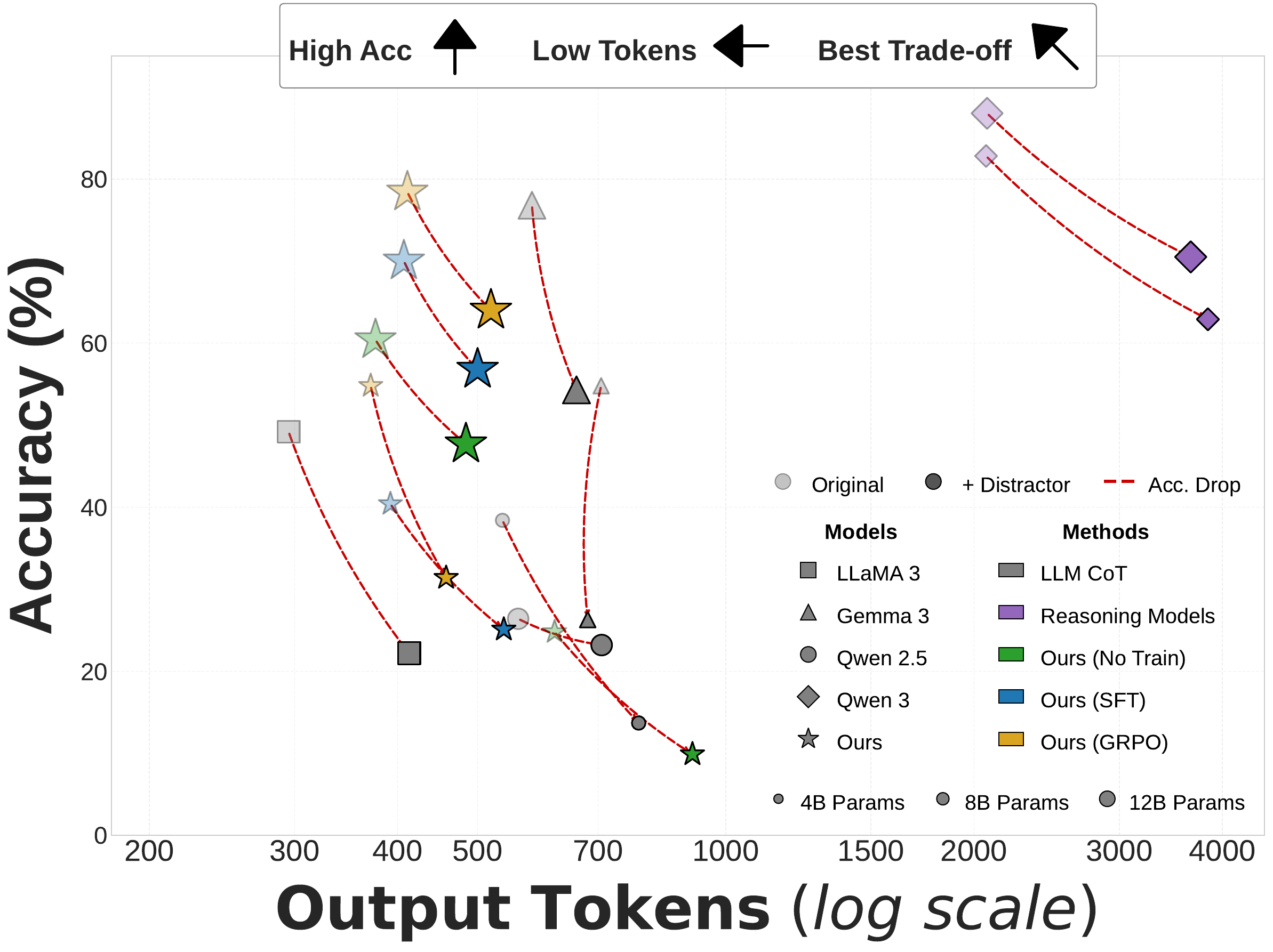}
   \caption{\textbf{MGSM}} 
 \end{subfigure}
 \hfill
 \begin{subfigure}[t]{0.32\textwidth}
   \includegraphics[width=\linewidth]{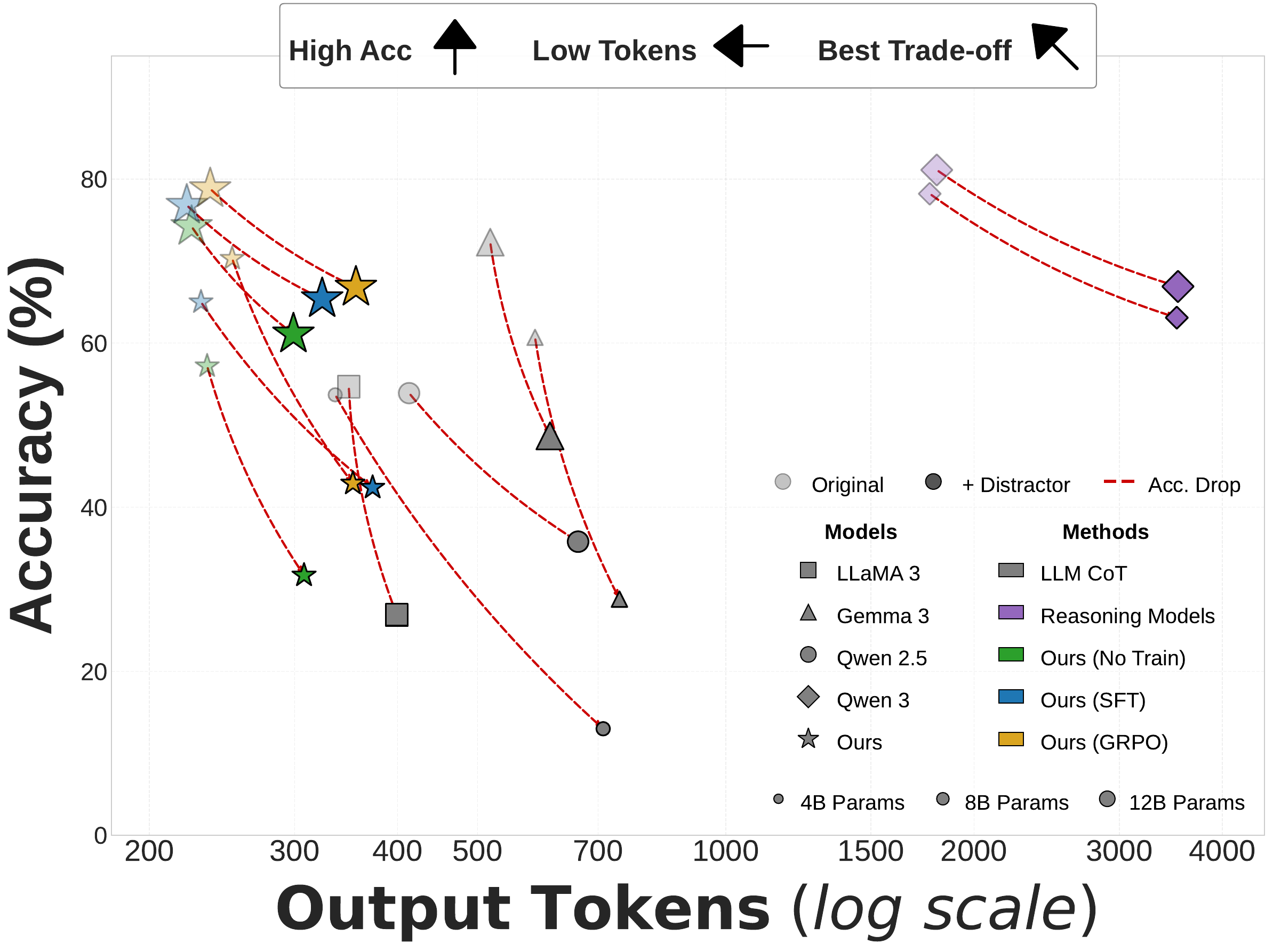}
   \caption{\textbf{MSVAMP}} 
 \end{subfigure}
 \hfill
 \begin{subfigure}[t]{0.32\textwidth}
   \includegraphics[width=\linewidth]{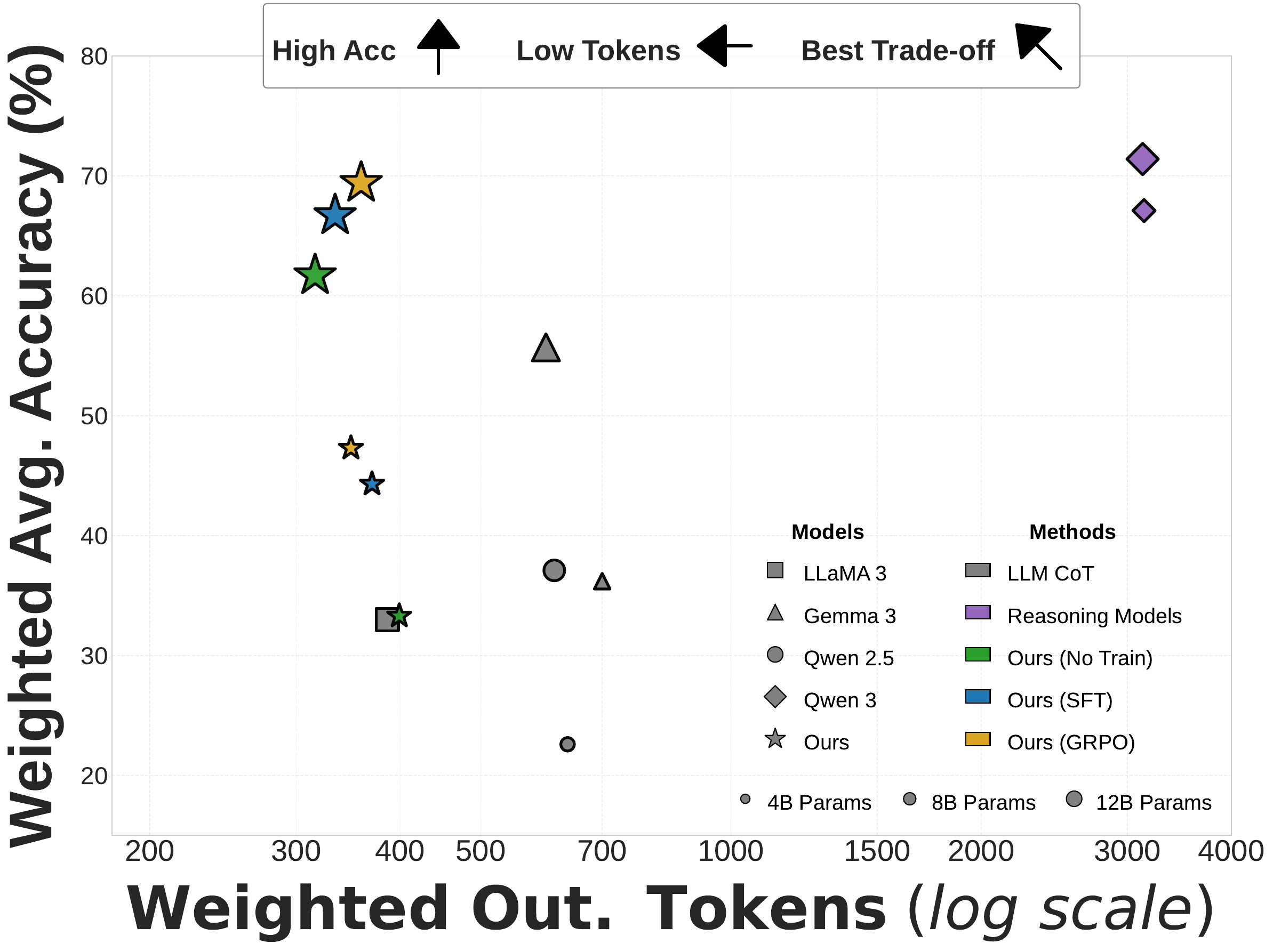}
   \caption{\textbf{Weighted Average}\textsuperscript{\protect\hyperref[app:weighted_avg_main]{*}}}
 \end{subfigure}
 \caption{\textbf{Accuracy vs. Efficiency.} We compare our method (\textcolor{ourgold}{\ensuremath{\bigstar}}) against CoT baselines and reasoning models. Dashed arrows in \textbf{(a)} and \textbf{(b)} show performance drops due to distractors. Marker size indicates model parameter count. Overall, our method approaches reasoning model accuracy with significantly lower computational cost.}
 \label{fig:all}
\end{figure*}


CoT prompting and code generation (PoT/PAL) have no 
mechanism in their output to distinguish relevant 
quantities from distractors - distractors can be 
silently incorporated without the model ever committing 
to their relevance (Section~\ref{sec:results}).
Reasoning-specialized models reduce this through 
extended computation, and their traces have been shown 
to encode implicit graph-like 
structures~\cite{minegishi2025topology}.
We propose making this structure explicit: the model 
constructs an executable directed acyclic graph that 
assigns each identified quantity a relevance annotation, 
followed by deterministic execution over only the 
relevant subgraph.

We reformulate mathematical word problem solving as a structured graph generation task. Given a problem statement $\mathcal{P}$, our approach instructs a language model to construct a directed acyclic graph $\mathcal{G} = (V, E)$ (prompt in \ref{app:graph_prompt}).
Each node $v$ is assigned a unique identifier and includes a boolean attribute $\texttt{distractor}(v)$ that indicates whether the node contributes to the solution path. Edges $e \in E$ encode data dependencies, where an edge from node $u$ to $v$ indicates that $v$ consumes the output of $u$ as input. The graph structure enforces a topological ordering that enables deterministic execution.

A distinguished node $v_{\text{final}} \in V$ is designated as the final result, representing the problem's solution. Graph execution proceeds through recursive computation: starting from $v_{\text{final}}$, we traverse backwards through dependencies, computing each node's value from its inputs until all constant nodes are reached \cite{geh2021learning}. This execution model provides verifiable correctness since each computational step is explicitly represented and can be independently validated (\ref{app:graph_exec}).

The graph representation offers several advantages over sequential reasoning chains. The explicit $\texttt{distractor}$ annotation enables the model to identify irrelevant numerical information during construction, preventing error propagation. The structured format separates problem understanding from computation, allowing errors to be localized to specific operations. Finally, the representation's token budget scales with problem complexity rather than explanation verbosity as shown in Section~\ref{sec:token_scaling}.

\subsection{Training Pipeline}

We leverage \texttt{GPT-4.1} as a teacher model to generate computational graph representations for training. For SFT, we curate 3K examples: 1.5K from the SOMADHAN dataset and 1.5K from numina-math-cot-bn \cite{paul2025leveraging, numina_math_datasets, rahman2024numinamathcotbn} \textit{without any distractor augmentation}. Each problem is processed by \texttt{GPT-4.1} to generate structured graph outputs, which are subsequently executed using our graph interpreter (Algorithm~\ref{app:graph_exec}
in Apps.) to verify computational correctness. Only verified samples are retained for training. For GRPO, we use 3K questions from numina-math-cot-bn 
without pre-generated solutions, allowing the model 
to explore multiple solution paths during training. 
The GRPO reward function $R(g,y)$ for a generated 
graph $g$ and ground truth $y$ is defined as
\[
R(g,y)=0.5\,\mathbb{I}_{\mathrm{fmt}}
+0.5\,\mathbb{I}_{\mathrm{exec}}
+\mathbb{I}_{\mathrm{acc}}\!\left(\mathrm{exec}(g),y\right),
\]
where \( \mathbb{I}_{\mathrm{fmt}} \), \( \mathbb{I}_{\mathrm{exec}} \), and \( \mathbb{I}_{\mathrm{acc}} \) are indicator functions corresponding to (i) a valid JSON schema with at least three nodes\footnote{This constraint prevents models from taking shortcuts by answering the question with a single node.}, (ii) successful topological execution, and (iii) an exact numerical match between the executed graph output and \( y \), respectively.
A natural concern with GRPO training is reward
hacking \citep{guo2025deepseek, dipta2026pa3, dipta2026decomposerl} - whether the model can achieve high
reward without genuine reasoning.
Our reward design mitigates this: 
$\mathbb{I}_{\mathrm{acc}}$ is computed by 
\emph{executing} the generated graph, so the model 
cannot reach the correct answer without first 
producing a structurally valid and executable 
computational path.
A model that satisfies format and execution checks 
but produces incorrect answers receives at most 
$0.5 + 0.5 = 1.0$ out of a maximum reward of $2.0$, 
providing limited learning signal without genuine 
reasoning improvement. We verify that GRPO does not collapse graphs 
into trivial structures through reward hacking in 
Appendix~\ref{app:reward-analysis}. 

We structure training as SFT followed by GRPO 
because they address complementary bottlenecks.
SFT requires ground-truth graph annotations from 
a teacher model, limiting 
scale to 3K examples, but teaches the model to 
produce well-formed executable graphs.
GRPO needs only final numerical answers, allowing 
exploration of diverse solution paths through 
multi-sample generation - but converges poorly
without a strong initialization \citep{dipta2026ganitllmdifficultyawarebengalimathematical}.
SFT therefore provides the structural initialization 
that allows GRPO to focus on reasoning quality - an 
effect confirmed by the ablation in 
Appendix~\ref{app:ablation_grpo_init}, where SFT$\to$GRPO 
consistently outperforms Base$\to$GRPO. 

\section{Experimental Setup}
\label{sec:setup}

\paragraph{Models}
We evaluate three model categories on
\ourdb:
\textbf{(1)}~standard LLMs with 5-shot CoT prompting:
Qwen~2.5 (3B/7B), LLaMA~3 8B, Gemma~3 (4B/12B);
\textbf{(2)}~reasoning-specialized models with extended thinking
budgets (up to 7{,}168 output tokens):
Qwen~3 (4B/8B);
\textbf{(3)}~\dag\textsc{Dagger}-trained Gemma-3 (4B/12B) using
SFT followed by GRPO.
Training details are in Apps.~\ref{app:train_config}.

\paragraph{Supplementary Evaluation}
To assess generality beyond fine-tuned models and Bangla,
we additionally evaluate GPT-4.1 in few-shot settings with
four prompting strategies (CoT \citep{wei2022chain}, PoT \citep{chen2022program}, GoT \citep{besta2024graph},
\dag\textsc{Dagger}) on distractor-augmented Bangla data,
and with CoT and \dag\textsc{Dagger} on original
MGSM for Telugu, Thai, English, and Spanish (Sec. \ref{sec:crosslingual}).

\paragraph{Metrics}
We report accuracy on each evaluation split,
accuracy drop ($\Delta$ = original $-$ augmented)
as a robustness measure, and average output tokens
as an efficiency measure.
We compute weighted average accuracy across all four
Bangla splits following dataset-size weighting
(Apps.~\ref{app:weighted_avg_main}).
For \dag\textsc{Dagger}, we additionally report
execution rate (fraction of generated graphs that
execute successfully). Failed executions are considered incorrect.

\section{Results and Analysis}
\label{sec:results}

\subsection{Distractor Impact on CoT and Reasoning Models}
\label{sec:rq1}

Table~\ref{tab:merged_results} presents the full evaluation on
\textsc{DistractMath-BN}.
We organize analysis around two findings.

\begin{figure}[t]
\centering
\includegraphics[width=0.95\columnwidth]{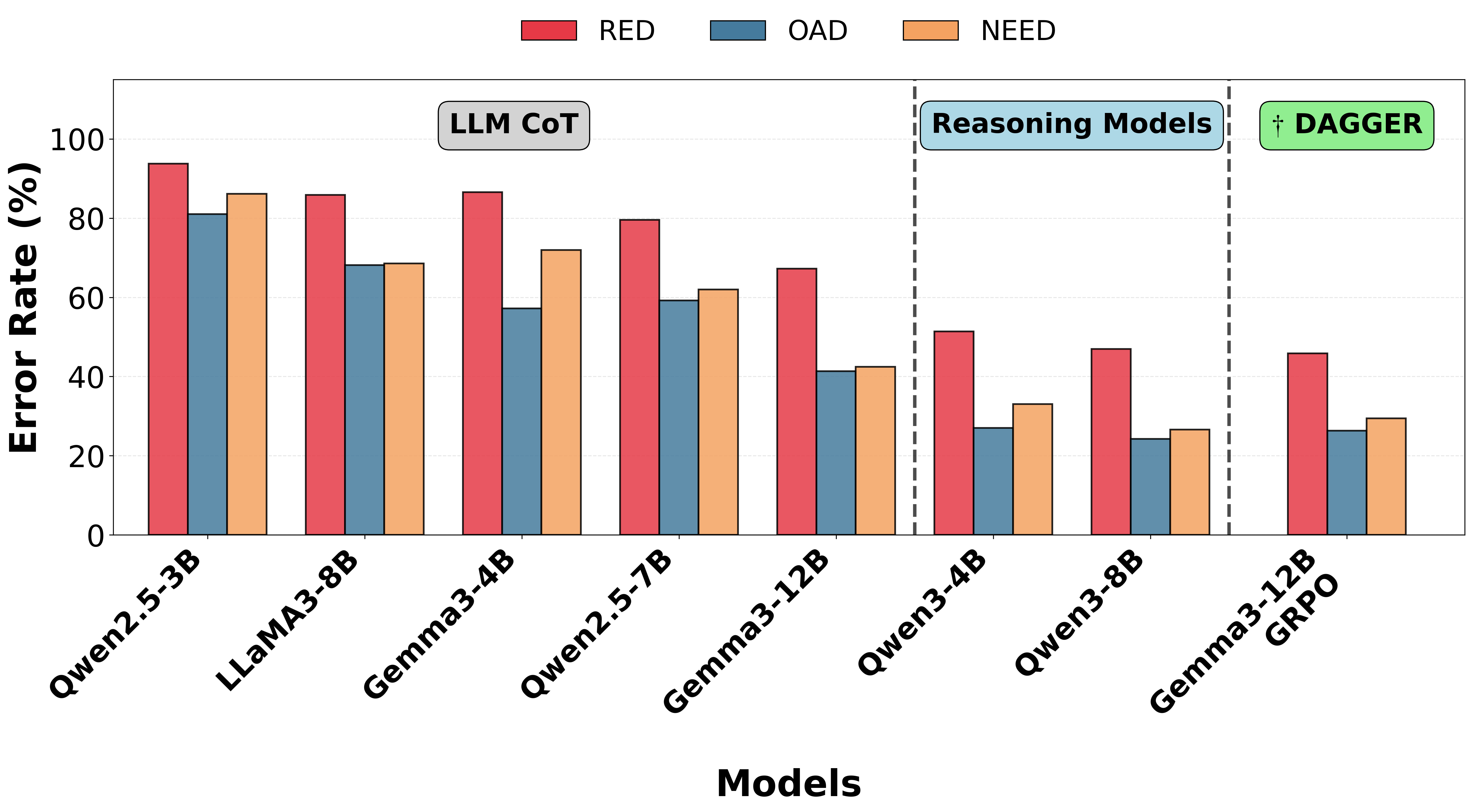}
\caption{Error rates across distractor types for different model categories.}
\label{fig:error_analysis}
\end{figure}

\paragraph{CoT Prompting Fails Catastrophically}
Standard 5-shot CoT prompting shows severe degradation
under distractors: accuracy drops up to
40.7 points, with Qwen~2.5 (3B) falling from 53.7\%
to 13.0\% on MSVAMP. LLaMA 3 8B shows a notably small drop of 3.2 points on MGSM, likely attributable to its already low baseline (26.4\%).
Figure~\ref{fig:error_analysis} reveals vulnerabilities
across all distractor categories.
Related Entity Distractors (RED) are most disruptive
(46--94\% error rate), indicating persistent
difficulties in entity-level disambiguation:
models cannot reliably distinguish quantities about
the focal entity from those about peripheral entities.
Orthogonal Attribute (24--81\%) and Null-Effect
(27--86\%) distractors also cause substantial errors,
confirming that the vulnerability is systematic rather
than category-specific.

\paragraph{Reasoning Models: Robust but Expensive}
Reasoning-specialized models perform substantially better:
Qwen~3-8B achieves 71.4\% weighted accuracy, the highest
among all evaluated models.
However, they still degrade 14.2--19.9 points and
consume $\sim$5.2$\times$ more tokens than standard
models (3{,}128 vs.\ $\sim$600 tokens per problem).
This establishes a key tension: achieving distractor
robustness through extended reasoning comes at
significant computational cost, motivating the
search for more efficient alternatives, particularly for low-resource languages.

\subsection{\dag\textsc{Dagger}: Executable Graph Reasoning for Efficiency}
\label{sec:rq2}

\dag\textsc{Dagger} addresses the robustness--efficiency
tension identified above.
Figure~\ref{fig:all} shows that Gemma-3 12B with
SFT$\rightarrow$GRPO achieves 69.4\% weighted accuracy
using only 359 tokens per problem.

\paragraph{Robustness}
The 12B GRPO model drops only 12.0--14.4 points under
distractors, compared to 14.2--17.5 for Qwen~3-8B
and up to 40.7 for standard CoT models.
This improved robustness emerges from the structured
representation: the computational graph requires
the model to commit to specific nodes and dependencies,
rather than processing all tokens uniformly.
Importantly, \dag\textsc{Dagger} is not trained on
distractor-augmented examples
(Appendix~\ref{app:distractor_dist}): training data
averages only 0.21 naturally present distractor nodes versus 2.7 in
evaluation, suggesting that the graph structure itself
provides an inductive bias for distractor filtering.

\paragraph{Efficiency}
\dag\textsc{Dagger} uses 89\% fewer tokens than reasoning
models (359 vs.\ 3{,}128) by encoding mathematical
relationships as graph edges rather than verbose
natural language explanations.

\paragraph{Accuracy--Efficiency Trade-off}
We note that Qwen~3-8B achieves higher absolute accuracy
(71.4\% vs.\ 69.4\%).
However, Qwen~3 is a reasoning-specialized model that
underwent extensive RL training specifically for
mathematical reasoning~\citep{yang2025qwen3} and
relies on extended test-time computation.
Our approach applies only 3K SFT + 3K GRPO examples
to Gemma-3, a general-purpose model.
The small accuracy gap comes with $\sim$9$\times$ fewer
tokens and better robustness, representing a practical
trade-off for resource-constrained deployment.
\begin{table}[t]
\centering
\small
\begin{tabular}{lcccc}
\toprule
& \multicolumn{2}{c}{\textbf{4B}} & \multicolumn{2}{c}{\textbf{12B}} \\
\cmidrule(lr){2-3}\cmidrule(lr){4-5}
\textbf{Annot.} & \textbf{Orig.} & \textbf{+Dist.} & \textbf{Orig.} & \textbf{+Dist.} \\
\midrule
w/o & 64.9 & 39.5 & 76.2 & 60.4 \\
w/ & 65.0 & 42.4 & 76.8 & 65.4 \\
\bottomrule
\end{tabular}
\caption{Ablation: effect of distractor node annotations 
on MSVAMP accuracy (\%). Annotations provide minimal 
impact on clean data but yield +2.9 (4B) and +5.0 (12B) 
gains under distractor augmentation.}
\label{tab:ablation-distractor}
\end{table}

\paragraph{Capacity Threshold}
Graph-based reasoning requires sufficient model 
capacity to be effective.
Without any training, \dag\textsc{Dagger} achieves 
33.3\% weighted accuracy at 4B but 61.7\% at 
12B - the latter exceeding all the 
CoT baselines - indicating that larger models can 
leverage the graph structure even in a few-shot prompting setup.
With SFT and GRPO, the 4B model improves to 47.3\% 
but still lags substantially behind the 12B model 
(69.4\%), suggesting a practical lower bound on
model size for constructing well-formed
computational graphs, mirroring the capacity floor
reported for Bangla math reasoning at smaller
scales \citep{dipta2026ganitllmdifficultyawarebengalimathematical}.

\paragraph{Effect of Distractor Annotations}
To isolate the contribution of explicit distractor 
node annotations beyond graph structure alone, we perform ablation experiment by
training models using SFT with identical setup but remove all 
distractor annotations from the graph schema 
(Table~\ref{tab:ablation-distractor}).

On clean data, distractor annotations have minimal 
effect (+0.1/+0.6), as expected since original problems 
contain few distractors.
Under distractor augmentation, annotations yield 
clear gains: +2.9 for 4B and +5.0 for 12B.
The benefit increases with model scale, suggesting 
larger models better leverage the annotation signal.
We note that consistent trends in 
Table~\ref{tab:merged_results} between SFT and GRPO 
suggest a similar pattern would hold after policy 
optimization.

\subsection{Comparison with Structured Reasoning}
\label{sec:structured-comp}
To contextualize \dag\textsc{Dagger} against alternative
structured approaches, we evaluate GPT-4.1 with CoT,
PoT, GoT, and \dag\textsc{Dagger} on distractor-augmented
\textsc{DistractMath-BN}
(Table~\ref{tab:structured}).

\begin{table}[t]
\centering
\small
\begin{tabular}{@{}lcccc}
\toprule
& \textbf{CoT} & \textbf{PoT} & \textbf{GoT} & \textbf{\dag D.} \\
\midrule
\rowcolor{gray!15}
\multicolumn{5}{l}{\textsc{DistractMath-BN} (MGSM)}\\
Accuracy & 64.9 & 56.0 & 66.8 & \textbf{69.2} \\
Exec.\ Rate & --- & 94.8 & --- & 98.4 \\
\midrule
\rowcolor{gray!15}
\multicolumn{5}{l}{\textsc{DistractMath-BN} (MSVAMP)} \\
Accuracy & 63.2 & 49.6 & 58.6 & \textbf{65.2} \\
Exec.\ Rate & --- & 96.4 & --- & 99.7 \\
\bottomrule
\end{tabular}
\caption{GPT-4.1 results on distractor-augmented \textsc{DistractMath-BN}.}
\label{tab:structured}
\end{table}

\dag\textsc{Dagger} achieves the highest accuracy on
both datasets. Notably, GoT requires $\sim$6 API calls 
per problem while all other methods use a single call, 
yet still underperforms \dag\textsc{Dagger} by 
2.4 and 6.6 points on MGSM and MSVAMP respectively.
PoT underperforms significantly despite high execution
rates (94--96\%), confirming that successful code
execution does not guarantee correct reasoning under
distractors \citep{xu2026patternaware}: sequential code generation, like free-form
CoT, processes all input tokens uniformly.
GoT's multi-call approach provides some robustness
through exploration of alternative paths, but at
substantially higher cost and without explicit
distractor filtering.

\subsection{Low-Resource Language Results}
\label{sec:crosslingual}

To assess whether \dag\textsc{Dagger}'s benefits extend
beyond Bangla and fine-tuned models, we evaluate GPT-4.1
on MGSM across four languages
(Table~\ref{tab:crosslingual}).

\begin{figure*}[t]
    \centering
    \hfill
    \begin{subfigure}[t]{0.44\textwidth}
        \centering
        \includegraphics[width=\linewidth]{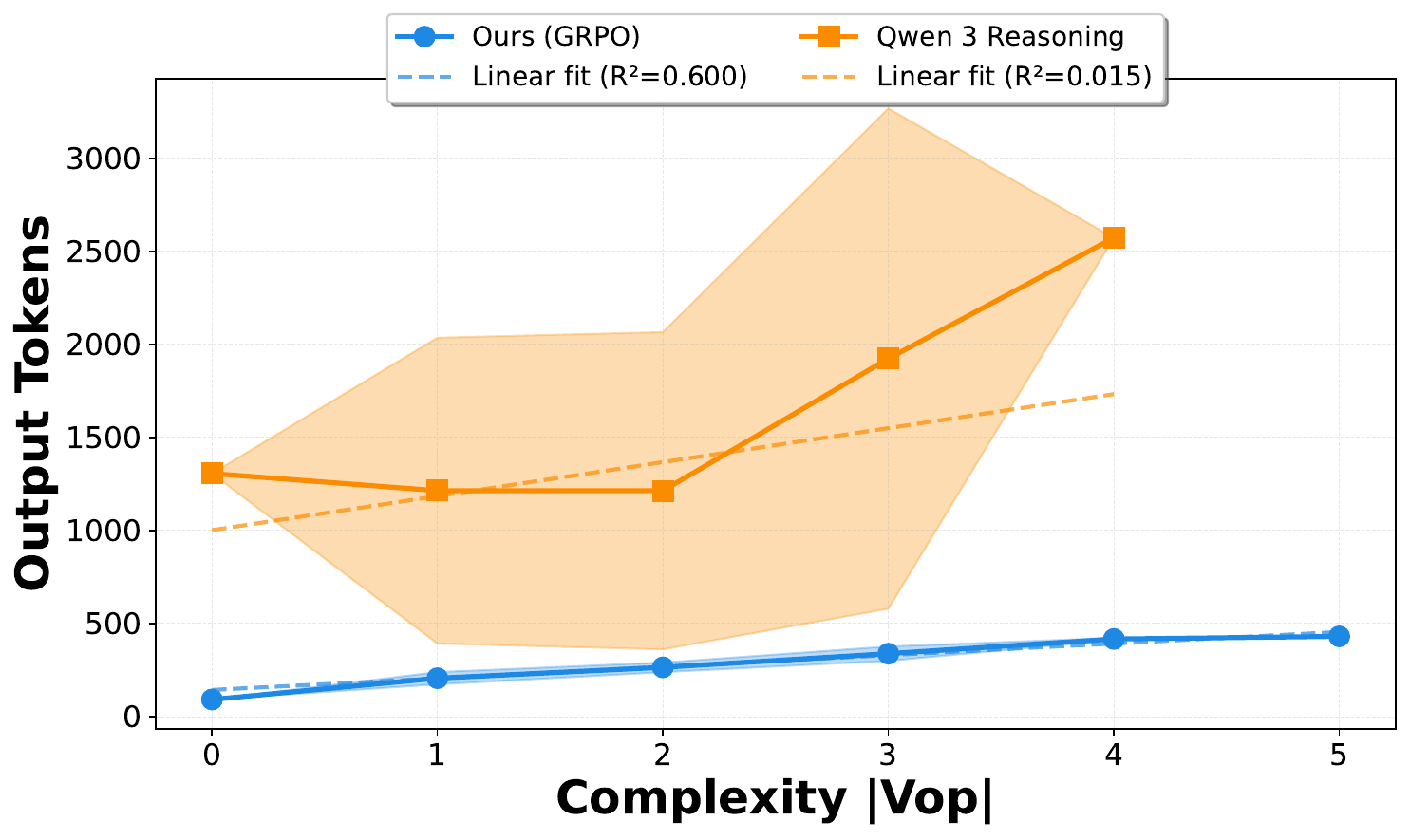}
        \caption{MSVAMP-BN (original)}
        \label{fig:token_scaling_msvamp}
    \end{subfigure}
    \hfill
    \begin{subfigure}[t]{0.44\textwidth}
        \centering
        \includegraphics[width=\linewidth]{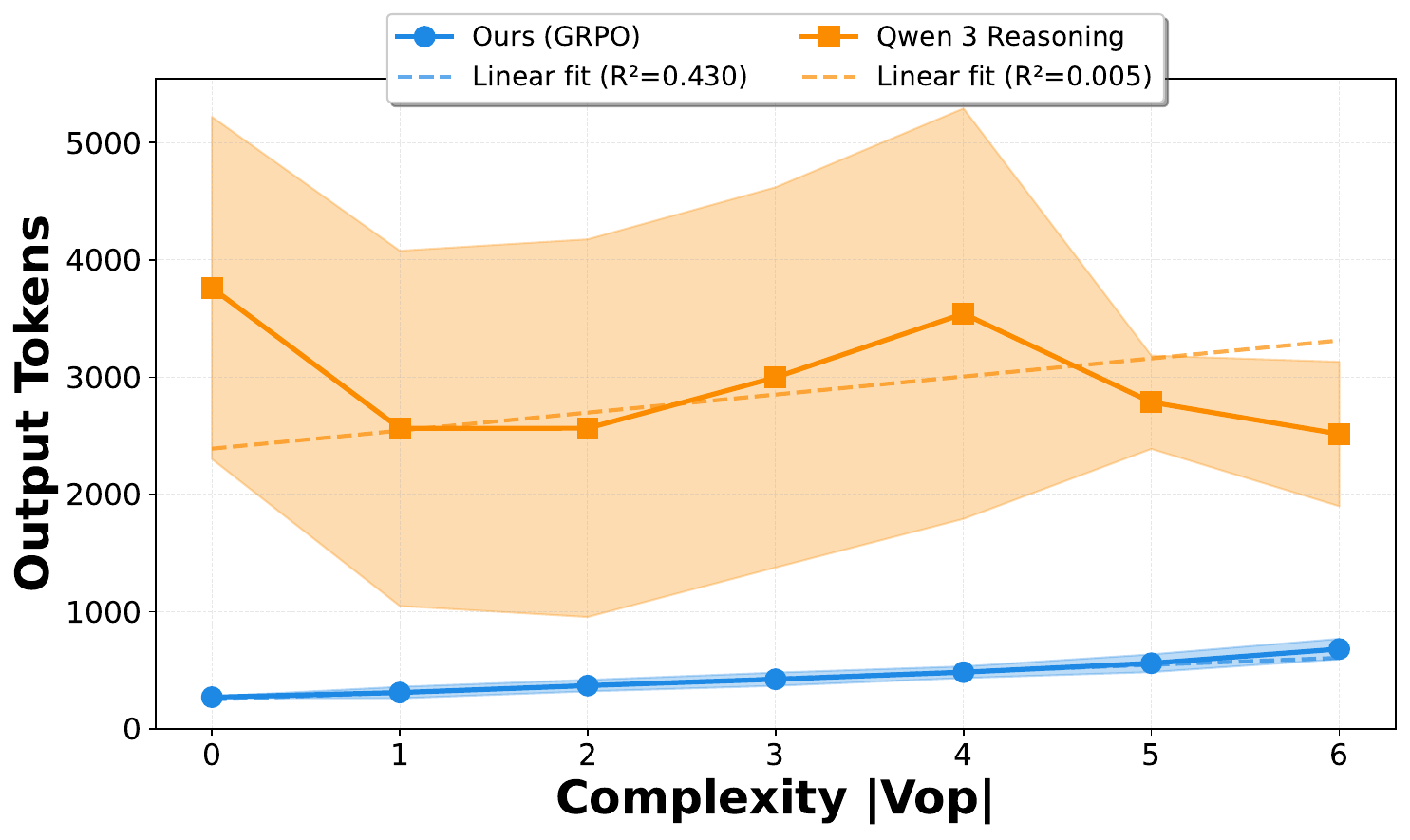}
        \caption{MSVAMP-BN + distractors (\ourdb)}
        \label{fig:token_scaling_msvamp_aug}
    \end{subfigure}
    \hfill
    \caption{Output token usage as a function of computational complexity ($|V_{\text{op}}|$).}
    \label{fig:token_scaling}
\end{figure*}

\dag\textsc{Dagger} consistently shows strong performance compared to CoT across
all four languages, with improvements ranging from
+4.4 on Telugu to +10.8 on English.
High execution rates ($>$93\%) across all languages
indicate that graph-based representations maintain
structural validity across diverse linguistic contexts. However, the CoT results are sensitive to prompt choice: Table~\ref{tab:crosslingual} reports the original MGSM 5-shot ICL setup (78.4\%), while alternate configurations obtain 80.0--89.6\% accuracy (Appendix~\ref{app:cot_prompt_sensitivity}). Thus, Table~\ref{tab:crosslingual} represents a fixed-prompt comparison.

These results suggest that the structured
graph representation generalizes beyond Bangla, though
the primary contribution of this work remains the
systematic evaluation of distractor robustness and
efficiency in the Bangla setting.

\begin{table}[t]
\centering
\small
\begin{tabular}{lcc}
\toprule
\textbf{Language} & \textbf{CoT (5-shot)} & \textbf{\ours (3-shot)} \\
\midrule
English & 78.4 & \textbf{89.2} \\
Spanish & 75.2 & \textbf{85.6} \\
Telugu & 74.4 & \textbf{78.8} \\
Thai & 78.4 & \textbf{87.6} \\
\bottomrule
\end{tabular}
\caption{GPT-4.1 accuracy (\%) on MGSM across four languages.
Execution rates for \dag\textsc{Dagger} exceed 93\%
in all cases.}
\label{tab:crosslingual}
\end{table}

\subsection{Token Scaling with Computational Complexity}
\label{sec:token_scaling}

We investigate whether output length in \ours scales with problem complexity rather than explanation verbosity. We define complexity as $|V_{\text{op}}|$, the number of operation nodes (excluding \texttt{const} and \texttt{identity}) in the ground-truth computational graph produced by the teacher model. This directly reflects how many arithmetic steps are needed to solve a problem.

We compare \ours (Gemma-3-12B, SFT$\rightarrow$GRPO) against Qwen-3-8B Reasoning on MSVAMP-BN, both on original and distractor-augmented splits. Each baseline output is assigned the $|V_{\text{op}}|$ of its corresponding ground-truth graph. The analysis is restricted to correctly solved instances (Figures~\ref{fig:token_scaling_msvamp} and~\ref{fig:token_scaling_msvamp_aug}).

On the original split, \ours output length correlates well with $|V_{\text{op}}|$ ($R^2{=}0.600$), staying in the hundreds of tokens with low variance. Qwen-3-8B shows almost no correlation ($R^2{=}0.015$), spanning thousands of tokens regardless of complexity. Under distractor augmentation, the pattern persists: \ours maintains meaningful correlation ($R^2{=}0.430$) while the reasoning baseline remains near-zero ($R^2{=}0.005$). These results show that graph-output length is associated with the number of operation nodes in the same generated graph, whereas free-form reasoning length is largely independent of problem complexity.

\section{Discussion}

Standard CoT prompting is brittle under distractors, dropping up to 40.7 points on Bangla MWPs across all distractor categories (Figure~\ref{fig:error_analysis}). Reasoning models reduce this gap but still degrade 14--20 points at ${\sim}$9$\times$ the token cost.

\dag\textsc{Dagger} addresses this by forcing explicit dependency commitments through the DAG structure, preventing distractors from entering the solution path, while distractor annotations allow acknowledging irrelevant quantities without incorporating them. This robustness emerges without distractor-augmented training (0.21 distractor nodes in training vs.\ 2.7 at evaluation), suggesting the graph structure itself induces filtering. Output length tracks computational complexity ($R^2{=}0.600$ vs.\ $R^2{=}0.015$ for free-form reasoning), yielding 89\% fewer tokens at comparable accuracy.

Related Entity Distractors remain hardest across all models (Figure~\ref{fig:error_analysis}), suggesting entity-level disambiguation may require richer representations beyond node-level annotations.

\section{Conclusion}

We introduce \ourdb, revealing that standard CoT prompting fails catastrophically under distractors while reasoning-specialized models achieve robustness at prohibitive computational cost. \ours demonstrates that structured computational graph generation enables comparable accuracy with substantially improved token efficiency. Critically, this robustness emerges without explicit distractor training. Our findings suggest structured representations provide a practical pathway for robust mathematical reasoning in noisy low-resource settings, though entity-level semantic disambiguation remains an open challenge.

\section*{Limitations}

Our systematic distractor generation employs JSON-structured outputs to ensure semantic consistency, defining distractors operationally as numerical values not reached during backward graph traversal. However, our benchmark is limited to arithmetic word problems in Bangla with only three distractor categories (Related Entity, Orthogonal Attribute, and Null-Effect Event) and does not address non-numeric distractors such as irrelevant textual descriptions. The quality assurance process relied on review by a single native Bangla-speaking PhD researcher, which may introduce individual judgment biases.

Our computational graph framework assumes problems can be decomposed into directed acyclic graphs with atomic operations, which may not extend to geometric reasoning, algebraic manipulation, or mathematical proof. It is designed for deterministic solutions and cannot represent probabilistic reasoning or problems with multiple valid solution paths. The complexity of generating well-formed graphs creates practical capacity requirements, suggesting an empirical lower bound on model size for effective performance, though smaller models still show improvements over baselines.

Our approach demonstrates benefits even without supervised fine-tuning, though combining SFT with GRPO yields optimal performance. The SFT phase creates a dependency on constructing high-quality initial training examples, which may limit scalability to resource-intensive annotation scenarios. While we demonstrate improvements for frontier models like GPT-4.1 in few-shot settings, our fine-tuning experiments are limited to the Gemma model family and may require adaptation for other architectures. The cross-lingual evaluation on four languages used limited test sets without distractor augmentation, providing preliminary evidence for transferability. Our evaluation relies on exact numeric match, which may not capture reasoning quality or penalize minor arithmetic errors proportionately. Finally, this work focuses on elementary arithmetic word problems and should not be generalized to advanced mathematical reasoning including algebra, geometry, or calculus.

\section*{Ethics Statement}

This work addresses the gap in NLP capabilities for Bangla, a widely spoken but underrepresented language in language technology. We have made \ourdb, code, and prompts publicly available. Dataset construction relies in part on a proprietary large language model, which may introduce biases or stylistic artifacts specific to that model and should be considered when interpreting results or extending the benchmark. Our method is designed to improve mathematical reasoning in educational technology and low-resource language support tools. While the structured representation improves interpretability, responsible deployment requires human oversight, particularly in educational settings.

\bibliography{custom}

\appendix
\clearpage

\section{Appendix}
\label{sec:appendix}

\noindent\textbf{Table of Contents}
{\small
\begin{description}[leftmargin=0em, itemsep=1pt, parsep=0pt, font=\normalfont]
    \item[\ref{app:weighted_avg_main}] Impact of Distractor Augmentation on Different Models
    \item[\ref{app:ablation_grpo_init}] Ablation: Effect of GRPO Initialization (Base vs.\ SFT)
    \item[\ref{app:cot_prompt_sensitivity}] CoT Prompt Sensitivity
    \item[\ref{app:reward-analysis}] Reward Structure Analysis
    \item[\ref{app:inference_config}] Inference Configuration
    \item[\ref{app:train_config}] Training Configuration
    \item[\hspace{1.5em}\ref{app:sft_config}] Supervised Fine-Tuning
    \item[\hspace{1.5em}\ref{app:grpo_config}] Group Relative Policy Optimization
    \item[\ref{app:distractor_dist}] Distractor Distribution Analysis
    \item[\ref{app:graph_exec}] Graph Execution Algorithm
    \item[\ref{app:qual_analysis}] Qualitative Analysis: \ours vs. Chain-of-Thought Reasoning
    \item[\hspace{1.5em}\ref{app:qual_example1}] Example 1
    \item[\hspace{1.5em}\ref{app:qual_example2}] Example 2
    \item[\hspace{1.5em}\ref{app:qual_example3}] Example 3
    \item[\ref{app:prompts}] Prompts
    \item[\hspace{1.5em}\ref{app:red_prompt}] Generating Related Entity Distractor
    \item[\hspace{1.5em}\ref{app:oad_prompt}] Generating Orthogonal Attribute Distractor
    \item[\hspace{1.5em}\ref{app:need_prompt}] Generating Null-Effect Event Distractor
    \item[\hspace{1.5em}\ref{app:gpt_verify}] Verifying Modified Problems
    \item[\hspace{1.5em}\ref{app:graph_prompt}] Generating Computational Graph
    \item[\hspace{1.5em}\ref{app:prompt_reasoning}] Prompt for Reasoning Models
    \item[\hspace{1.5em}\ref{app:cot}] Prompt for Chain of Thought (5-Shot)
    \item[\ref{app:examples}] Examples of Augmented Questions from \ourdb
\end{description}
}

\subsection{Impact of Distractor Augmentation on Different Models}
\label{app:weighted_avg_main}

Table~\ref{tab:merged_results} reports the full tabular results corresponding to the trends summarized in Figure~\ref{fig:all}.
Beyond the aggregate view in Figure~\ref{fig:all}, the table provides per-dataset performance under both original and distractor-augmented settings, together with accuracy drops and output-token statistics.

We compute a weighted average accuracy over four evaluation splits using dataset-size weighting:

{\scriptsize
\[
\mathrm{Acc}_{\mathrm{w}}
=
\frac{
n_{\mathrm{MG}}A_{\mathrm{MG}}
+n_{\mathrm{MS}}A_{\mathrm{MS}}
+n_{\mathrm{MG}_{+D}}A_{\mathrm{MG}_{+D}}
+n_{\mathrm{MS}_{+D}}A_{\mathrm{MS}_{+D}}
}{
n_{\mathrm{MG}}+n_{\mathrm{MS}}+n_{\mathrm{MG}_{+D}}+n_{\mathrm{MS}_{+D}}
}.
\]
}

Here, $\mathrm{MG}$ and $\mathrm{MS}$ denote the original MGSM and MSVAMP splits, while $\mathrm{MG}_{+D}$ and $\mathrm{MS}_{+D}$ denote their distractor-augmented counterparts.
$A_{\cdot}$ represents accuracy on each split, and $n_{\cdot}$ denotes the corresponding number of problems, with
$n_{\mathrm{MG}}{=}250$, $n_{\mathrm{MS}}{=}1000$, $n_{\mathrm{MG}_{+D}}{=}738$, and $n_{\mathrm{MS}_{+D}}{=}2947$.

\subsection{Ablation: Effect of GRPO Initialization (Base vs.\ SFT)}
\label{app:ablation_grpo_init}

\paragraph{Motivation.}
While Group Relative Policy Optimization (GRPO) enforces correctness through executable and verifiable rewards, it remains unclear whether GRPO alone is sufficient to induce structured computational graph generation, or whether an intermediate Supervised Fine-Tuning (SFT) phase is necessary to bootstrap this behavior.
This ablation isolates the contribution of \emph{initialization} by comparing GRPO trained directly from the original pre-trained checkpoint (\emph{Base$\rightarrow$GRPO}) against GRPO initialized from an SFT checkpoint (\emph{SFT$\rightarrow$GRPO}).

\paragraph{Setup.}
We train two GRPO variants for Gemma-3-4B and Gemma-3-12B using identical GRPO hyperparameters, reward definition, and training data.
The only difference is the initialization:
(i) \textbf{Base$\rightarrow$GRPO} starts from the original pre-trained Gemma-3 checkpoint, while
(ii) \textbf{SFT$\rightarrow$GRPO} starts from the SFT checkpoint trained on verified computational graph demonstrations.
We evaluate both variants on MGSM and MSVAMP (Bangla), as well as their distractor-augmented counterparts from \ourdb.

\paragraph{Results.}
Table~\ref{tab:ablation_grpo_base_vs_sft} summarizes the results.
Across both model sizes and datasets, GRPO initialized from SFT consistently outperforms GRPO trained directly from the base model, with the gap being substantially larger under distractor augmentation.

\begin{table}[t]
\centering
\scriptsize
\setlength{\tabcolsep}{4pt}
\renewcommand{\arraystretch}{1.15}
\begin{tabular}{lcccc}
\toprule
\textbf{Model} &
\textbf{MGSM} &
\textbf{MSVAMP} &
\textbf{MGSM (+D)} &
\textbf{MSVAMP (+D)} \\
\midrule
\multicolumn{5}{c}{\textbf{Gemma-3-4B}} \\
\midrule
Base$\rightarrow$GRPO & 29.2 & 57.1 & 13.1 & 29.3 \\
SFT$\rightarrow$GRPO  & 54.8 & 70.3 & 31.4 & 42.9 \\
\midrule
\multicolumn{5}{c}{\textbf{Gemma-3-12B}} \\
\midrule
Base$\rightarrow$GRPO & 67.6 & 75.0 & 48.4 & 59.6 \\
SFT$\rightarrow$GRPO  & 78.4 & 78.8 & 64.0 & 66.8 \\
\bottomrule
\end{tabular}
\vspace{1mm}
\caption{\textbf{Ablation on GRPO initialization.}
Accuracy (\%) on Bangla MGSM and MSVAMP, with and without distractor augmentation (+D).
SFT initialization consistently improves GRPO performance, especially under distractors.}
\label{tab:ablation_grpo_base_vs_sft}
\vspace{7mm}
\end{table}

For Gemma-3-12B, Base$\rightarrow$GRPO already achieves reasonable performance on clean data (67.6\% on MGSM, 75.0\% on MSVAMP), but degrades sharply under distractors (48.4\% and 59.6\%).
Initializing GRPO from SFT yields consistent gains across all settings, with particularly large improvements on augmented MGSM (+15.6 points) and augmented MSVAMP (+7.2 points), indicating improved robustness to irrelevant context.

For Gemma-3-4B, the effect of initialization is even more pronounced.
Base$\rightarrow$GRPO struggles to learn reliable graph generation, achieving only 29.2\% on MGSM and collapsing to 13.1\% on augmented MGSM.
In contrast, SFT$\rightarrow$GRPO nearly doubles accuracy on MGSM (54.8\%) and more than doubles performance under distractors (31.4\% on augmented MGSM).
This suggests that smaller models require SFT to learn basic structural and execution constraints before GRPO can effectively refine reasoning strategies.

Overall, these results show that while GRPO can improve performance even from a base checkpoint, \emph{SFT provides a critical scaffold} that stabilizes learning and enables GRPO to focus on semantic parsing and distractor filtering rather than merely discovering valid executable structures.
This explains why SFT$\rightarrow$GRPO yields the strongest robustness and accuracy throughout our experiments.

\subsection{CoT Prompt Sensitivity}
\label{app:cot_prompt_sensitivity}

To assess sensitivity to in-context examples and instruction format, we evaluate GPT-4.1 on English MGSM using four CoT prompting configurations - three different ICL settings and one without ICL \cite{xu2026mgsm}. Table~\ref{tab:cot_prompt_sensitivity} shows substantial variation across prompts.

\begin{table}[h]
\centering
\small
\setlength{\tabcolsep}{6pt}
\begin{tabular}{lc}
\toprule
\textbf{Configuration} & \textbf{Accuracy (\%)} \\
\midrule
5-shot CoT, original MGSM ICL set & 78.4 \\
5-shot CoT, alternate MGSM ICL set & 80.0 \\
3-shot CoT, alternate MGSM ICL set & 82.4 \\
MGSM-Pro instruction prompt (no ICL) & \textbf{89.6} \\
\bottomrule
\end{tabular}
\caption{Prompt sensitivity of GPT-4.1 CoT on English MGSM.}
\label{tab:cot_prompt_sensitivity}
\end{table}

\subsection{Reward Structure Analysis}
\label{app:reward-analysis}

We analyze graph complexity and reward component 
distributions across training phases to assess 
whether GRPO training produces structurally 
meaningful improvements.

\paragraph{Graph Complexity Across Training Phases}
Table~\ref{tab:graph-complexity} reports the average 
number of nodes per generated graph at each stage 
of training, evaluated on MSVAMP using Gemma-3 12B.
Average node count increases from 7.33 (without 
training) to 7.74 (SFT) to 7.84 (GRPO), 
indicating that GRPO does not reduce graph 
complexity relative to SFT.

\begin{table}[h]
\centering
\small
\begin{tabular}{lc}
\toprule
\textbf{Phase} & \textbf{Avg.\ Nodes} \\
\midrule
Without training & 7.33 \\
After SFT & 7.74 \\
After GRPO & 7.84 \\
\bottomrule
\end{tabular}
\caption{Average number of nodes per generated graph 
across training phases (Gemma-3 12B, MSVAMP).}
\label{tab:graph-complexity}
\end{table}

\paragraph{Reward Component Breakdown}
After SFT, format compliance 
($\mathbb{I}_{\mathrm{fmt}}$) and execution success 
($\mathbb{I}_{\mathrm{exec}}$) are already high, 
leaving limited room for improvement on these 
components.
The observed gains from GRPO (+2.7 weighted points 
for 12B, +3.0 for 4B) are concentrated in 
$\mathbb{I}_{\mathrm{acc}}$: the model produces 
graphs that execute to the correct answer more 
frequently, while format and execution scores 
remain largely unchanged.

\paragraph{Reward Hacking Prevention}
The compositional reward design imposes several 
structural constraints that limit reward exploitation.
Format compliance requires valid JSON with at least 
3 nodes, ruling out degenerate single-operation graphs.
Execution success requires all node dependencies to 
resolve during topological traversal - referencing 
a nonexistent dependency causes execution failure 
and zero execution reward.
The accuracy reward evaluates the numerical output 
of graph execution against ground truth, not the 
graph structure itself: a graph that is well-formatted 
and executes successfully but computes the wrong 
result receives zero accuracy reward.
Combined with the empirical evidence above - stable 
graph complexity and gains concentrated in 
accuracy - these constraints prevent reward hacking and improve genuine reasoning and distractor identification capabilities.

\subsection{Inference Configuration}
\label{app:inference_config}
For all chain-of-thought (CoT) baselines, we use the default inference configuration provided by the underlying model implementation.
Models are loaded in full precision without quantization, using BF16 arithmetic and a maximum sequence length of 4096 tokens.
During inference, we allow up to 2048 output tokens to accommodate multi-step reasoning traces.
Unless otherwise specified, all remaining decoding and runtime hyperparameters are kept at their default values.

For reasoning-specialized models, we employ a larger context window to support longer reasoning traces.
Specifically, models are loaded with a maximum sequence length of 8192 tokens in BF16 precision.
To prevent premature truncation of extended reasoning chains, we allow up to 7168 output tokens during generation.
All other inference-time settings follow the default configuration of the respective models.

For DAGGER-style inference without training, we use the same default inference setup as above but restrict the maximum number of generated output tokens to 1024.
\subsection{Training Configuration}
\label{app:train_config}
\subsubsection{Supervised Fine-Tuning} \label{app:sft_config}

We fine-tune the base Gemma-3 models (4B and 12B) using Rank-Stabilized LoRA (RSLoRA) with rank 64. The training uses a global batch size of 256 across 4 epochs with AdamW optimizer. We employ cosine learning rate scheduling starting from $1 \times 10^{-5}$ with minimum rate of $1 \times 10^{-6}$. Gradient checkpointing and sequence packing are enabled to optimize memory usage. Training is performed in BF16 precision with maximum gradient norm clipping at 1.0.

\vspace{2mm}

\begin{table}[h]
\centering
\scriptsize
\begin{tabular}{ll}
\toprule
\textbf{Parameter} & \textbf{Value} \\
\midrule
Base Model & Gemma-3 (4B/12B) \\
LoRA Rank / Alpha & 64 / 128 \\
Max Seq Length & 4096 \\
Global Batch Size & 256 \\
Epochs & 4 \\
Learning Rate & $10^{-5} \rightarrow 10^{-6}$ \\
Optimizer & AdamW \\
Weight Decay & 0.001 \\
Precision & BF16 \\
\bottomrule
\end{tabular}
\caption{SFT hyperparameters.}
\end{table}

\vspace{3mm}

\begin{table}[h]
\centering
\scriptsize
\begin{tabular}{ll}
\toprule
\textbf{Parameter} & \textbf{Value} \\
\midrule
Base Model & SFT checkpoint \\
LoRA Rank / Alpha & 64 / 128 \\
Max Prompt / Completion & 2048 / 2048 \\
Global Batch Size & 32 \\
Generations per Prompt & 8 \\
Epochs & 4 \\
Learning Rate & $10^{-5} \rightarrow 10^{-6}$ \\
Loss Type & BNPO \\
$\beta$ / $\epsilon$ / $\epsilon_{\text{high}}$ & 0.0 / 0.2 / 0.28 \\
Sampling (top-p / top-k) & 0.95 / 64 \\
Temperature & 1.0 \\
Precision & BF16 \\
\bottomrule
\end{tabular}
\caption{GRPO hyperparameters.}
\end{table}

\subsubsection{Group Relative Policy Optimization} \label{app:grpo_config}

We train the SFT checkpoints using GRPO with a structured reward function that assigns +0.5 for valid JSON graph format, +0.5 for successful execution, and +1.0 for correct answers (maximum 2.0 per generation). We employ GRPO with BNPO loss as it is specifically designed for tasks with verifiable outputs and discrete reward signals, making it well-suited for computational graph generation where correctness can be deterministically evaluated. The group-based policy optimization allows efficient comparison across multiple generated graphs per prompt, enabling the model to learn from relative quality differences. BNPO's binary normalization is particularly effective for our structured three-tier reward system (format validation, execution success, and answer correctness). For each prompt, we generate 8 completions using VLLM with top-p sampling (0.95) and top-k (64). The training uses BNPO loss with $\beta=0.0$, epsilon clipping at 0.2-0.28, and temperature 1.0. We employ LoRA rank 64 with global batch size 32 over 4 epochs. The learning rate follows cosine scheduling from $1 \times 10^{-5}$ to $1 \times 10^{-6}$ with 20 warmup steps.

\vspace{2mm}

\subsection{Distractor Distribution Analysis}
\label{app:distractor_dist}

To better understand the robustness gap between training and evaluation, we analyze the distribution of distractor nodes across the training and augmented evaluation datasets.
Table~\ref{tab:distractor_stats} summarizes the average, minimum, and maximum number of distractors per problem.

\begin{table}[h]
\centering
\scriptsize
\setlength{\tabcolsep}{6pt}
\renewcommand{\arraystretch}{1.15}
\begin{tabular}{lccc}
\toprule
\textbf{Dataset} & \textbf{Avg.\ Distractors} & \textbf{Min} & \textbf{Max} \\
\midrule
MGSM (+Distractor)     & 2.76 & 2 & 5 \\
MSVAMP (+Distractor)  & 2.68 & 2 & 4 \\
SFT Training Data     & 0.21 & 0 & 10 \\
\bottomrule
\end{tabular}
\caption{Distribution of distractor nodes across training and augmented evaluation datasets.}
\label{tab:distractor_stats}
\end{table}

As shown in Table~\ref{tab:distractor_stats}, the augmented MGSM and MSVAMP evaluation sets contain substantially more distractors per problem than the data used for supervised fine-tuning.
This imbalance arises because no explicit distractor augmentation is applied during SFT; instead, training samples are randomly drawn from SOMADHAN and Numina Math CoT datasets and no distractor augmentation is performed, hence naturally most examples contain few or no distractor nodes.
As a result, the average number of distractors in training remains very low despite a small number of outliers.

Importantly, although the model is not trained on augmented examples, it still demonstrates improved robustness under distractor-heavy evaluation.
This suggests that the structured design of distractor nodes in the computational graph formulation itself provides a strong inductive bias for robustness.
We expect that explicitly applying distractor augmentation during training would further improve performance; however, these results confirm that meaningful robustness gains can already be achieved through graph-level distractor design alone, even without direct exposure to augmented data.

\newpage
\subsection{Graph Execution Algorithm}
\label{app:graph_exec}
We present the high-level pseudo-code for executing the computational graph. The algorithm recursively computes node values while handling distractors through topological traversal.

\vspace{3mm}

\begin{tcolorbox}[
  enhanced,
  colback=blue!5,
  colframe=blue!60!black,
  boxrule=1pt,
  arc=2mm,
  left=1mm,
  right=1mm,
  top=3mm,
  bottom=3mm,
  breakable,
  title={\textbf{Algorithm: Computational Graph Execution}},
  fonttitle=\bfseries\sffamily\small,
  coltitle=white,
  attach boxed title to top center={yshift=-2mm},
  boxed title style={
    colback=blue!60!black,
    arc=2mm,
  },
]

\small

\noindent\textbf{Input:} Graph JSON: $G = \{V, E\}$ where $V$ are nodes

\noindent\textbf{Output:} Final numeric result $r \in \mathbb{R}$

\vspace{3mm}

\begin{tcolorbox}[
  colback=white,
  colframe=blue!30,
  boxrule=0.5pt,
  arc=1mm,
  left=2mm,
  right=2mm,
  top=2mm,
  bottom=2mm,
]
\textbf{\textsc{Function}} \texttt{ExecuteGraph}$(G)$:

\vspace{1mm}
\hspace{5mm} $\mathcal{N} \gets$ \textsc{Parse}$(G)$ \hfill \fcolorbox{gray!40}{gray!10}{\scriptsize\textcolor{gray!70}{Parse \& validate}}

\hspace{5mm} $\textit{cache} \gets \emptyset$, \hspace{2mm} $\textit{visited} \gets \emptyset$ \hfill \fcolorbox{gray!40}{gray!10}{\scriptsize\textcolor{gray!70}{Initialize}}

\hspace{5mm} $v_{\textit{final}} \gets$ \textsc{FindFinalNode}$(\mathcal{N})$

\hspace{5mm} \textbf{return} \textsc{Compute}$(v_{\textit{final}}, \mathcal{N}, \textit{cache}, \textit{visited})$
\end{tcolorbox}

\vspace{2mm}

\begin{tcolorbox}[
  colback=white,
  colframe=blue!30,
  boxrule=0.5pt,
  arc=1mm,
  left=2mm,
  right=2mm,
  top=2mm,
  bottom=2mm,
  breakable,
]
\textbf{\textsc{Function}} \texttt{Compute}$(v, \mathcal{N}, \textit{cache}, \textit{visited})$:

\vspace{1mm}
\hspace{5mm} \textbf{if} $v \in \textit{cache}$:

\hspace{10mm} \textbf{return} $\textit{cache}[v]$ \hfill \fcolorbox{blue!40}{blue!10}{\scriptsize\textcolor{blue!70}{Memoized}}

\vspace{1mm}
\hspace{5mm} \textbf{if} $v \in \textit{visited}$:

\hspace{10mm} \textbf{raise} \textsc{CircularDependency} \hfill \fcolorbox{red!40}{red!10}{\scriptsize\textcolor{red!70}{Cycle detect}}

\vspace{1mm}
\hspace{5mm} $\textit{visited} \gets \textit{visited} \cup \{v\}$

\hspace{5mm} $\textit{node} \gets \mathcal{N}[v]$

\hspace{5mm} $\textit{op} \gets \textit{node}.\texttt{op}$

\vspace{2mm}
\hspace{5mm} \textcolor{blue!70}{\textit{// Execute operation}}

\hspace{5mm} \textbf{if} $\textit{op} = \texttt{const}$:

\hspace{10mm} $r \gets \textit{node}.\texttt{val}$

\vspace{1mm}
\hspace{5mm} \textbf{else if} $\textit{op} = \texttt{add}$:

\hspace{10mm} $r \gets \sum_{a \in \textit{node}.\texttt{args}}$ \textsc{Resolve}$(a, ...)$

\vspace{1mm}
\hspace{5mm} \textbf{else if} $\textit{op} = \texttt{sub}$:

\hspace{10mm} $r \gets$ \textsc{Resolve}$(\textit{node}.\texttt{args}[0], ...) - $ \textsc{Resolve}$(\textit{node}.\texttt{args}[1], ...)$

\vspace{1mm}
\hspace{5mm} \textbf{else if} $\textit{op} = \texttt{mul}$:

\hspace{10mm} $r \gets \prod_{a \in \textit{node}.\texttt{args}}$ \textsc{Resolve}$(a, ...)$

\vspace{1mm}
\hspace{5mm} \textbf{else if} $\textit{op} = \texttt{div}$:

\hspace{10mm} $r \gets$ \textsc{Resolve}$(\textit{node}.\texttt{args}[0], ...) / $ \textsc{Resolve}$(\textit{node}.\texttt{args}[1], ...)$

\vspace{1mm}
\hspace{5mm} \textbf{else if} $\textit{op} \in \{\texttt{floor}, \texttt{ceil}, \texttt{round}, ...\}$:

\hspace{10mm} $r \gets$ \textsc{ApplyOp}$(\textit{op}, \textit{node}.\texttt{args}, ...)$

\vspace{2mm}
\hspace{5mm} $\textit{cache}[v] \gets r$ \hfill \fcolorbox{blue!40}{blue!10}{\scriptsize\textcolor{blue!70}{Cache result}}

\hspace{5mm} $\textit{visited} \gets \textit{visited} \setminus \{v\}$

\hspace{5mm} \textbf{return} $r$
\end{tcolorbox}

\vspace{2mm}

\begin{tcolorbox}[
  colback=white,
  colframe=blue!30,
  boxrule=0.5pt,
  arc=1mm,
  left=2mm,
  right=2mm,
  top=2mm,
  bottom=2mm,
]
\textbf{\textsc{Function}} \texttt{Resolve}$(a, \mathcal{N}, \textit{cache}, \textit{visited})$:

\vspace{1mm}
\hspace{5mm} \textbf{if} $a \in \mathbb{R}$:

\hspace{10mm} \textbf{return} $a$ \hfill \fcolorbox{green!40}{green!10}{\scriptsize\textcolor{green!70}{Numeric literal}}

\vspace{1mm}
\hspace{5mm} \textbf{else if} $a \in \mathcal{N}$:

\hspace{10mm} \textbf{return} \textsc{Compute}$(a, \mathcal{N}, \textit{cache}, \textit{visited})$ \hfill \fcolorbox{green!40}{green!10}{\scriptsize\textcolor{green!70}{Node ref}}

\vspace{1mm}
\hspace{5mm} \textbf{else}:

\hspace{10mm} \textbf{raise} \textsc{InvalidArgument}
\end{tcolorbox}

\end{tcolorbox}

\vspace{2mm}

\noindent\textbf{Complexity:} $\mathcal{O}(|V| + |E|)$ where $|V|$ is the number of nodes and $|E|$ is the number of edges.

\subsection{Qualitative Analysis: \ours vs. Chain-of-Thought Reasoning} \label{app:qual_analysis}

We present qualitative examples demonstrating how \ours correctly identifies and filters distractor information mitigating the limitations of chain-of-thought based approaches for Bangla MWP.

\vspace{3mm}

\subsubsection{Example 1} \label{app:qual_example1}

\noindent\textbf{Dataset:} MGSM-BN 

\noindent\textbf{Augmentation Type:} Related Entity Distractor (RED)

\vspace{2mm}

\begin{tcolorbox}[
  colback=gray!5,
  colframe=gray!40,
  boxrule=0.5pt,
  arc=1mm,
  left=2mm,
  right=2mm,
  top=3mm,
  bottom=3mm,
]
\textbf{Question:}

{\bengalifont জিনের কাছে 30টি ললিপপ আছে। \textcolor{red}{দোকানে আরও ৫০টি ললিপপ বিক্রি হছিল।} \textcolor{red}{তার ছোটবোনের কাছে ১৮টি ললিপপ আছে।} জিন 2টি ললিপপ খেয়েছে। \textcolor{red}{জিনের বন্ধু মিমি প্রতিদিন ৩টি ললিপপ খায়।} অবশিষ্ট ললিপপগুলি দিয়ে, জিন একটি ব্যাগের মধ্যে 2টি করে ললিপপ ভরতে চায়। তাহলে জিন কতগুলি ব্যাগ ভর্তি করতে পারবে?}

\vspace{2mm}
\textit{Jean has 30 lollipops. \textcolor{red}{50 more lollipops were being sold at the store.} \textcolor{red}{Her younger sister has 18 lollipops.} Jean ate 2 lollipops. \textcolor{red}{Jean's friend Mimi eats 3 lollipops per day.} With the remaining lollipops, Jean wants to fill bags with 2 lollipops each. How many bags can Jean fill?}

\vspace{2mm}
\textbf{Ground Truth Answer: 14}
\end{tcolorbox}

\vspace{3mm}

\noindent\textbf{Gemma 3 12B (CoT)} \hfill Predicted Answer: \textcolor{red}{48}

\vspace{2mm}

\begin{tcolorbox}[
  colback=white,
  colframe=red!60,
  boxrule=1pt,
  arc=1mm,
  left=2mm,
  right=2mm,
  top=3mm,
  bottom=3mm,
  breakable,
]
{\small\bengalifont 
প্রথমে, জিন 30টি ললিপপ কিনেছে এবং \textcolor{red}{তার ছোটবোনের কাছে 18টি ললিপপ ছিল}। জিন 2টি ললিপপ খেয়েছে। তার মানে জিন \textcolor{red}{30 + 18 - 2 = 46টি} ললিপপ নিয়ে শুরু করেছে। এরপর \textcolor{red}{জিন 50টি ললিপপ কিনেছে}। তার মানে জিন \textcolor{red}{46 + 50 = 96টি} ললিপপ নিয়ে শুরু করেছে। মিমি প্রতিদিন 3টি করে ললিপপ খায়। এই তথ্যটি এখানে অপ্রাসঙ্গিক। জিন 2টি করে ললিপপ ভরতে চায়। তার মানে জিন \textcolor{red}{96 / 2 = 48টি} ব্যাগ ভর্তি করতে পারবে। উত্তর হল 48।
}

\vspace{2mm}
\textbf{Error Analysis:} Incorrectly includes sister's lollipops (18) and store lollipops (50) in Jean's count, leading to wrong calculation: 30 + 18 + 50 - 2 = 96, then 96 / 2 = 48.
\end{tcolorbox}

\vspace{3mm}

\noindent\textbf{Qwen 3 8B (Reasoning)} \hfill Predicted Answer: \textcolor{red}{23}

\vspace{2mm}

\begin{tcolorbox}[
  colback=white,
  colframe=red!60,
  boxrule=1pt,
  arc=1mm,
  left=2mm,
  right=2mm,
  top=3mm,
  bottom=3mm,
  breakable,
]
{\scriptsize
\texttt{<think>} Okay, let's try to figure out how many bags Jean can fill with the lollipops. Hmm, first, I need to find out the total number of lollipops Jean has. Let me start by adding up all the lollipops from different sources.

So, Jean has 30 lollipops. Then, there are 50 more lollipops being sold at the store. Wait, does that mean those 50 are added to Jean's collection? Or are they separate? The problem says ``{\normalfont\bengalifont দোকানে আরও ৫০টি ললিপপ বিক্রি হচ্ছিল}'' which translates to ``there are 50 more lollipops being sold at the store.'' ... \textcolor{red}{The problem might be that Jean has 30, and then there are 50 more at the store, and her sister has 18}. But the question is about Jean's lollipops. So maybe the 50 at the store are not hers. ...

... \textcolor{red}{So maybe the total lollipops Jean has are 30 (hers) + 18 (sister's) + 50 (store's)?} But that seems odd. Unless Jean is combining all of them. But the problem doesn't specify that. Alternatively, maybe the 50 at the store are not hers, and the sister's 18 are hers. \textcolor{red}{So Jean has 30 + 18 = 48? Then she ate 2, so 48 - 2 = 46}. Then Mimi eats 3 per day, but how many days? Wait, the problem says ``{\normalfont\bengalifont জিনের বন্ধু মিমি প্রতিদিন ৩টি ললিপপ খায়}'' (Jean's friend Mimi eats 3 lollipops per day). But how many days? The problem doesn't mention the number of days. ...

... Maybe the problem is that the 50 at the store are being sold, so Jean might have bought them? But the problem doesn't say she bought them. Alternatively, maybe the total lollipops Jean has are 30 (hers) + 18 (sister's) = 48. Then she ate 2, so 46. Then Mimi eats 3 per day, but again, no days. ...

... Given the ambiguity, I think the most logical assumption is that the 50 at the store are not part of Jean's lollipops, and the sister's 18 are hers. So total lollipops: 30 + 18 = 48. Jean ate 2, so 46. Then Mimi eats 3 per day, but since the problem doesn't specify the number of days, maybe it's a mistake and we should ignore Mimi's consumption. \textcolor{red}{So 46 divided by 2 is 23 bags}. But this is also a guess. ... So the answer is 23. \texttt{</think>}
}

\vspace{2mm}
\textbf{Error Analysis:} Confused by distractor information; incorrectly adds sister's lollipops (18) to Jean's count, resulting in (30 + 18 - 2) / 2 = 23.
\end{tcolorbox}

\vspace{3mm}

\noindent\textbf{\ours (Ours)} \hfill Predicted Answer: \textcolor{green!60!black}{14}

\vspace{2mm}

\begin{tcolorbox}[
  colback=white,
  colframe=green!60,
  boxrule=1pt,
  arc=1mm,
  left=2mm,
  right=2mm,
  top=3mm,
  bottom=3mm,
  breakable,
]
{\scriptsize\ttfamily
\{\\
~~"nodes": [\\
~~~~\{"id": "jin\_start", "op": "const", "val": 30, "distractor": false,\\
~~~~~~"label": "{\normalfont\bengalifont জিনের প্রাথমিক ললিপপ}"\},\\
~~~~\{"id": "store\_lollipop", "op": "const", "val": 50, \textcolor{green!60!black}{"distractor": true},\\
~~~~~~"label": "{\normalfont\bengalifont দোকানে বিক্রি হওয়া ললিপপ}"\},\\
~~~~\{"id": "sister\_lollipop", "op": "const", "val": 18, \textcolor{green!60!black}{"distractor": true},\\
~~~~~~"label": "{\normalfont\bengalifont ছোট বোনের ললিপপ}"\},\\
~~~~\{"id": "jin\_eaten", "op": "const", "val": 2, "distractor": false,\\
~~~~~~"label": "{\normalfont\bengalifont জিন খেয়েছে}"\},\\
~~~~\{"id": "jin\_left", "op": "sub", "args": ["jin\_start", "jin\_eaten"],\\
~~~~~~"distractor": false, "label": "{\normalfont\bengalifont জিনের অবশিষ্ট}"\},\\
~~~~\{"id": "mimmi\_eat", "op": "const", "val": 3, \textcolor{green!60!black}{"distractor": true},\\
~~~~~~"label": "{\normalfont\bengalifont মিমি প্রতিদিন খায়}"\},\\
~~~~\{"id": "bag\_capacity", "op": "const", "val": 2, "distractor": false\},\\
~~~~\{"id": "bags\_full", "op": "div", "args": ["jin\_left", "bag\_capacity"],\\
~~~~~~"distractor": false\},\\
~~~~\{"id": "full\_bags", "op": "floor", "args": ["bags\_full"],\\
~~~~~~"distractor": false\},\\
~~~~\{"id": "final\_result", "op": "identity", "args": ["full\_bags"],\\
~~~~~~"distractor": false, "label": "{\normalfont\bengalifont চূড়ান্ত উত্তর}"\}\\
~~]\\
\}
}

\vspace{2mm}
\textbf{Success:} Correctly identifies distractors (store=50, sister=18, Mimi=3) and computes only from Jean's lollipops: (30 - 2) / 2 = 14.
\end{tcolorbox}

\newpage
\subsubsection{Example 2} \label{app:qual_example2}

\noindent\textbf{Dataset:} MGSM-BN

\noindent\textbf{Augmentation Type:} Orthogonal Attribute Distractors (OAD)

\vspace{2mm}

\begin{tcolorbox}[
  colback=gray!5,
  colframe=gray!40,
  boxrule=0.5pt,
  arc=1mm,
  left=2mm,
  right=2mm,
  top=3mm,
  bottom=3mm,
]
\textbf{Question:}

{\bengalifont পাঞ্চো একদিনে 20 মাইল হাঁটেন। \textcolor{red}{প্রতিদিন হাঁটার জন্য তিনি ১.৫ ঘণ্টা সময় নেন।} সপ্তাহান্তে ব্যতিরেকে তিনি 10 মাইল হাঁটেন। \textcolor{red}{সপ্তাহান্তে হাঁটার সময় ৪৫ মিনিট লাগে।} এক সপ্তাহে তিনি কত মাইল হাঁটেন?}

\vspace{2mm}
\textit{Pancho walks 20 miles a day. \textcolor{red}{He takes 1.5 hours per day for walking.} Except for weekends, he walks 10 miles. \textcolor{red}{Walking on weekends takes 45 minutes.} How many miles does he walk in a week?}

\vspace{2mm}
\textbf{Ground Truth Answer: 120}
\end{tcolorbox}

\vspace{3mm}

\noindent\textbf{Gemma 3 12B (CoT)} \hfill Predicted Answer:  \textcolor{red}{110}

\vspace{2mm}

\begin{tcolorbox}[
  colback=white,
  colframe=red!60,
  boxrule=1pt,
  arc=1mm,
  left=2mm,
  right=2mm,
  top=3mm,
  bottom=3mm,
  breakable,
]
{\small\bengalifont 
সপ্তাহান্তে ব্যতিরেকে তিনি 10 মাইল হাঁটেন। তিনি 5 দিন ধরে 20 মাইল করে হাঁটেন। সুতরাং, তিনি 5 * 20 = 100 মাইল হাঁটেন। \textcolor{red}{সপ্তাহান্তে তিনি 10 মাইল হাঁটেন}। সুতরাং, তিনি \textcolor{red}{100 + 10 = 110} মাইল হাঁটেন। উত্তর হল 110।
}

\vspace{2mm}
\textbf{Error Analysis:} Correctly calculates weekday miles (5 × 20 = 100) but fails to multiply weekend miles by 2 days. Uses 10 instead of 20 (10 × 2), resulting in 110 instead of 120.
\end{tcolorbox}

\vspace{3mm}

\noindent\textbf{Qwen 3 8B (Reasoning)} \hfill Predicted Answer:  \textcolor{red}{N/A}

\vspace{2mm}

\begin{tcolorbox}[
  colback=white,
  colframe=red!60,
  boxrule=1pt,
  arc=1mm,
  left=2mm,
  right=2mm,
  top=3mm,
  bottom=3mm,
  breakable,
]
{\scriptsize
\texttt{<think>} Okay, let's try to figure out how many miles Pancho walks in a week. ... The first sentence says ``{\normalfont\bengalifont পাঞ্চো একদিনে 20 মাইল হাঁটেন।}'' So that translates to ``Pancho walks 20 miles a day.'' But then it says ``{\normalfont\bengalifont প্রতিদিন হাঁটার জন্য তিনি ১.৫ ঘণ্টা সময় নেন।}'' Which means ``He takes 1.5 hours per day for walking.'' ...

... Then it says: ``{\normalfont\bengalifont সপ্তাহান্তে ব্যতিরেকে তিনি 10 মাইল হাঁটেন।}'' Which translates to ``Except for weekends, he walks 10 miles.'' \textcolor{red}{Wait, that's confusing}. ... The phrase ``{\normalfont\bengalifont সপ্তাহান্তে ব্যতিরেকে}'' means ``except for weekends'' or ``excluding weekends.'' ... \textcolor{red}{So that would mean that on weekdays, he walks 10 miles, and on weekends, he walks something else? But the first sentence said he walks 20 miles a day. So there's a contradiction here}. ...

... Wait, maybe the problem is structured as follows: On weekdays, he walks 20 miles per day, taking 1.5 hours each day. On weekends, he walks 10 miles, taking 45 minutes. But then the third sentence says ``{\normalfont\bengalifont সপ্তাহান্তে ব্যতিরেকে তিনি 10 মাইল হাঁটেন।}'' Which is ``Except for weekends, he walks 10 miles.'' \textcolor{red}{So that would mean that on weekdays, he walks 10 miles, and on weekends, he walks something else. But the first sentence says he walks 20 miles a day. So this is conflicting}. ...

... I think the confusion comes from the translation of the third sentence. Let me try to parse it again: ``{\normalfont\bengalifont সপ্তাহান্তে ব্যতিরেকে তিনি 10 মাইল হাঁটেন।}'' ... So the sentence is saying ``Except for weekends, he walks 10 miles.'' \textcolor{red}{So that would mean that on weekdays, he walks 10 miles, and on weekends, he walks something else. But the first sentence says he walks 20 miles a day. So this is conflicting}. ...

... \textcolor{red}{I think I need to proceed with the assumption that the first sentence is saying that on weekdays, he walks 20 miles each day, and on weekends, he walks 10 miles each day}. Then, the third sentence says ``except for weekends, he walks 10 miles,'' which would mean that on weekdays, he walks 10 miles, and on ... \texttt{</think>}

\vspace{1mm}
{\scriptsize\textit{}}
}

\vspace{2mm}
\textbf{Error Analysis:} Gets extremely confused by the problem structure and distractor information (time), unable to parse the question correctly and loops in contradictory reasoning without reaching an answer even after generating 7168 tokens.
\end{tcolorbox}

\vspace{3mm}

\noindent\textbf{\ours (Ours)} \hfill Predicted Answer: \textcolor{green!60!black}{120}

\vspace{2mm}

\begin{tcolorbox}[
  colback=white,
  colframe=green!60,
  boxrule=1pt,
  arc=1mm,
  left=2mm,
  right=2mm,
  top=3mm,
  bottom=3mm,
  breakable,
]
{\scriptsize\ttfamily
\{\\
~~"nodes": [\\
~~~~\{"id": "weekday\_miles", "op": "const", "val": 20, "distractor": false,\\
~~~~~~"label": "{\normalfont\bengalifont প্রতিদিন কর্মদিবসে হাঁটা (মাইল)}"\},\\
~~~~\{"id": "weekday\_time", "op": "const", "val": 1.5, \textcolor{green!60!black}{"distractor": true},\\
~~~~~~"label": "{\normalfont\bengalifont প্রতিদিন কর্মদিবসে হাঁটার সময় (ঘণ্টা)}"\},\\
~~~~\{"id": "weekend\_miles", "op": "const", "val": 10, "distractor": false,\\
~~~~~~"label": "{\normalfont\bengalifont প্রতিদিন সপ্তাহান্তে হাঁটা (মাইল)}"\},\\
~~~~\{"id": "weekend\_time", "op": "const", "val": 0.75, \textcolor{green!60!black}{"distractor": true},\\
~~~~~~"label": "{\normalfont\bengalifont সপ্তাহান্তে হাঁটার সময় (ঘণ্টা, ৪৫ মিনিট)}"\},\\
~~~~\{"id": "weekdays", "op": "const", "val": 5, "distractor": false,\\
~~~~~~"label": "{\normalfont\bengalifont কর্মদিবস সংখ্যা}"\},\\
~~~~\{"id": "weekend\_days", "op": "const", "val": 2, "distractor": false,\\
~~~~~~"label": "{\normalfont\bengalifont সপ্তাহান্তের দিন সংখ্যা}"\},\\
~~~~\{"id": "total\_weekday\_miles", "op": "mul",\\
~~~~~~"args": ["weekday\_miles", "weekdays"], "distractor": false,\\
~~~~~~"label": "{\normalfont\bengalifont কর্মদিবসে মোট হাঁটা (মাইল)}"\},\\
~~~~\{"id": "total\_weekend\_miles", "op": "mul",\\
~~~~~~"args": ["weekend\_miles", "weekend\_days"], "distractor": false,\\
~~~~~~"label": "{\normalfont\bengalifont সপ্তাহান্তে মোট হাঁটা (মাইল)}"\},\\
~~~~\{"id": "total\_miles", "op": "add",\\
~~~~~~"args": ["total\_weekday\_miles", "total\_weekend\_miles"],\\
~~~~~~"distractor": false, "label": "{\normalfont\bengalifont এক সপ্তাহে মোট হাঁটা (মাইল)}"\},\\
~~~~\{"id": "final\_result", "op": "identity", "args": ["total\_miles"],\\
~~~~~~"distractor": false, "label": "{\normalfont\bengalifont চূড়ান্ত উত্তর}"\}\\
~~]\\
\}
}

\vspace{2mm}
\textbf{Success:} Correctly identifies time-related information as distractors (1.5 hours, 45 minutes) and computes: (20 × 5) + (10 × 2) = 100 + 20 = 120.
\end{tcolorbox}

\newpage
\subsubsection{Example 3} \label{app:qual_example3}

\noindent\textbf{Dataset:} MGSM-BN 

\noindent\textbf{Augmentation Type:} Null-Effect Event Distractors (NEED)

\vspace{2mm}

\begin{tcolorbox}[
  colback=gray!5,
  colframe=gray!40,
  boxrule=0.5pt,
  arc=1mm,
  left=2mm,
  right=2mm,
  top=3mm,
  bottom=3mm,
]
\textbf{Question:}

{\bengalifont লয়েডের একটি এগ ফার্ম আছে। \textcolor{red}{একদিন লয়েড ভেবেছিলেন ডিমের দাম বাড়াবেন, কিন্তু পরে আর বাড়ালেন না।} প্রতিদিন তাঁর মুরগিরা 252টি করে ডিম দেয় ও তিনি সেগুলি প্রতি ডজন 2 ডলার দরে বিক্রয় করেন। \textcolor{red}{এক সপ্তাহে তাঁর এক বন্ধু ৩ ডজন ডিম কিনতে চেয়েছিলেন, কিন্তু শেষে আর নেননি।} তিনি প্রতি সপ্তাহে ডিম থেকে কত উপার্জন করেন? \textcolor{red}{যদি লয়েড সপ্তাহে অতিরিক্ত ৫০টি ডিম পেতেন, তাহলে উপার্জন আরও বেশি হতো, কিন্তু তা হয়নি।}}

\vspace{2mm}
\textit{Lloyd has an egg farm. \textcolor{red}{One day Lloyd thought about raising egg prices, but later didn't.} Every day his hens lay 252 eggs and he sells them at \$2 per dozen. \textcolor{red}{In a week, his friend wanted to buy 3 dozen eggs but didn't buy them in the end.} How much does he earn per week from eggs? \textcolor{red}{If Lloyd had received an extra 50 eggs per week, his earnings would have been higher, but that didn't happen.}}

\vspace{2mm}
\textbf{Ground Truth Answer: 294}
\end{tcolorbox}

\vspace{3mm}

\noindent\textbf{Gemma 3 12B (CoT)} \hfill Predicted Answer:  \textcolor{red}{288}

\vspace{2mm}

\begin{tcolorbox}[
  colback=white,
  colframe=red!60,
  boxrule=1pt,
  arc=1mm,
  left=2mm,
  right=2mm,
  top=3mm,
  bottom=3mm,
  breakable,
]
{\small\bengalifont 
প্রথমে, আমাদের প্রতিদিনের ডিম বিক্রির সংখ্যা বের করতে হবে। যেহেতু তিনি প্রতিদিন 252টি ডিম পাড়েন এবং প্রতি ডজন ডিমের দাম 2, তাই প্রথমে ডিমের সংখ্যাকে ডজনে পরিবর্তন করতে হবে। 252টি ডিম / 12 ডিম/ডজন = 21 ডজন ডিম। এরপর, প্রতিদিনের আয় হবে 21 ডজন * 2/ডজন = 42। সুতরাং, প্রতিদিনের আয় 42। এরপর, প্রতি সপ্তাহে আয় হবে 42/দিন * 7 দিন/সপ্তাহ = 294। \textcolor{red}{তার বন্ধু ৩ ডজন ডিম কিনতে চেয়েছিল, কিন্তু কেনেনি। তাই ৩ ডজন ডিমের দাম বাদ দিতে হবে। ৩ ডজন ডিম * 2/ডজন = 6। সুতরাং, প্রতি সপ্তাহে লয়েডের আয় হবে 294 - 6 = 288।} উত্তর হল 288।
}

\vspace{2mm}
\textbf{Error Analysis:} Correctly calculates weekly earnings as \$294 but then incorrectly subtracts \$6 for the friend's eggs that were never purchased, resulting in \$288.
\end{tcolorbox}

\vspace{3mm}

\noindent\textbf{Qwen 3 8B (Reasoning)} \hfill Predicted Answer:  \textcolor{green!60!black}{294}

\vspace{2mm}

\begin{tcolorbox}[
  colback=white,
  colframe=green!60,
  boxrule=1pt,
  arc=1mm,
  left=2mm,
  right=2mm,
  top=3mm,
  bottom=3mm,
  breakable,
]
{\scriptsize
\texttt{<think>} Okay, let's try to figure out how much money Lloyd earns from selling eggs each week. ... The problem says he has a farm with hens. Each day, his hens lay 252 eggs. He sells them at \$2 per dozen. ...

... So in a week, which is 7 days, the total number of eggs would be 252 multiplied by 7. ... 252 times 7 is 1764 eggs per week. ... Now, since he sells them by the dozen, I need to convert the total number of eggs into dozens. One dozen is 12 eggs. So 1764 divided by 12. ... That's 147 dozens. ...

... Then, he sells each dozen for \$2. So 147 dozens times \$2 per dozen. ... So his weekly earnings would be \$294. ... The friend didn't end up buying them, it doesn't affect the total earnings because he still sells all the eggs he has. ... So the answer should be \$294. \texttt{</think>}

\vspace{1mm}
{\normalfont\bengalifont প্রশ্ন বিশ্লেষণ:}

{\normalfont\bengalifont লয়েডের আয়ের হিসাব করার জন্য আমাদের মোট ডিমের সংখ্যা বের করে সেটিকে ডজনে রূপান্তর করতে হবে এবং তারপর মোট মূল্য নির্ধারণ করতে হবে। প্রশ্নে কিছু বিভ্রান্তিকর তথ্য আছে, যেমন: বন্ধুর ডিম না কেনা বা কাল্পনিক অতিরিক্ত ৫০টি ডিম, যা মূল হিসাবে প্রভাব ফেলবে না।}

{\normalfont\bengalifont ধাপ দ্বারা ধাপ সমাধান:}

{\normalfont\bengalifont ১. মোট ডিমের সংখ্যা নির্ণয়: লয়েডের মুরগিরা প্রতিদিন ২৫২টি ডিম দেয়। এক সপ্তাহে (৭ দিন) মোট ডিমের সংখ্যা: $252 \times 7 = 1764$ টি ডিম।}

{\normalfont\bengalifont ২. ডজনে রূপান্তর: আমরা জানি, ১ ডজন = ১২টি ডিম। সুতরাং, ১৭৬৪টি ডিমের ডজন সংখ্যা: $\frac{1764}{12} = 147$ ডজন।}

{\normalfont\bengalifont ৩. মোট উপার্জন নির্ণয়: প্রতি ডজন ডিমের বিক্রয়মূল্য 2। সুতরাং, এক সপ্তাহে মোট উপার্জন: $147 \times 2 = 294$ ডলার।}

{\normalfont\bengalifont উত্তর: 294।}
}

\vspace{2mm}
\textbf{Success:} Correctly identifies that the friend's unpurchased eggs and hypothetical extra eggs are distractors that don't affect the actual earnings calculation.
\end{tcolorbox}

\vspace{3mm}

\noindent\textbf{\ours (Ours)} \hfill Predicted Answer:  \textcolor{green!60!black}{294}

\vspace{2mm}

\begin{tcolorbox}[
  colback=white,
  colframe=green!60,
  boxrule=1pt,
  arc=1mm,
  left=2mm,
  right=2mm,
  top=3mm,
  bottom=3mm,
  breakable,
]
{\scriptsize\ttfamily
\{\\
~~"nodes": [\\
~~~~\{"id": "eggs\_per\_day", "op": "const", "val": 252, "distractor": false,\\
~~~~~~"label": "{\normalfont\bengalifont প্রতিদিন ডিম উৎপাদন}"\},\\
~~~~\{"id": "days\_in\_week", "op": "const", "val": 7, "distractor": false,\\
~~~~~~"label": "{\normalfont\bengalifont এক সপ্তাহে দিন}"\},\\
~~~~\{"id": "total\_eggs\_week", "op": "mul",\\
~~~~~~"args": ["eggs\_per\_day", "days\_in\_week"], "distractor": false,\\
~~~~~~"label": "{\normalfont\bengalifont সপ্তাহে মোট ডিম}"\},\\
~~~~\{"id": "dozen", "op": "const", "val": 12, "distractor": false,\\
~~~~~~"label": "{\normalfont\bengalifont এক ডজন ডিম}"\},\\
~~~~\{"id": "price\_per\_dozen", "op": "const", "val": 2, "distractor": false,\\
~~~~~~"label": "{\normalfont\bengalifont প্রতি ডজন ডিমের দাম}"\},\\
~~~~\{"id": "dozens\_sold", "op": "div", "args": ["total\_eggs\_week", "dozen"],\\
~~~~~~"distractor": false, "label": "{\normalfont\bengalifont মোট ডজন ডিম}"\},\\
~~~~\{"id": "total\_earning", "op": "mul",\\
~~~~~~"args": ["dozens\_sold", "price\_per\_dozen"], "distractor": false,\\
~~~~~~"label": "{\normalfont\bengalifont সপ্তাহে মোট উপার্জন}"\},\\
~~~~\{"id": "friend\_eggs", "op": "const", "val": 3, \textcolor{green!60!black}{"distractor": true},\\
~~~~~~"label": "{\normalfont\bengalifont বন্ধুর ডিমের চাহিদা (ডজন)}"\},\\
~~~~\{"id": "extra\_eggs", "op": "const", "val": 50, \textcolor{green!60!black}{"distractor": true},\\
~~~~~~"label": "{\normalfont\bengalifont অতিরিক্ত ডিম (কিন্তু পায়নি)}"\},\\
~~~~\{"id": "final\_result", "op": "identity", "args": ["total\_earning"],\\
~~~~~~"distractor": false, "label": "{\normalfont\bengalifont চূড়ান্ত উত্তর}"\}\\
~~]\\
\}
}

\vspace{2mm}
\textbf{Success:} Correctly identifies negation-based distractors (friend's unpurchased 3 dozen eggs and hypothetical extra 50 eggs) and computes: $(252 \times 7) / 12 \times 2 = 147 \times 2 = 294$.
\end{tcolorbox}

\newpage
\subsection{Prompts} \label{app:prompts}

\subsubsection{Prompt for Generating Related Entity Distractor}
\label{app:red_prompt}
\begin{promptbox}[Related Entity Distractor Generation]
\small
\textbf{Your task:} Add 1-3 new \textbf{distractor sentences} to a Bengali math problem. These sentences add information about the item whose quantity is being asked for or related entities or intermediate facts, but DO NOT change the final answer.

\medskip
\textbf{RULES:}
\begin{enumerate}
\item You may mention the same type of item being asked about, but only when referring to other people, places, or unrelated situations - you must NOT change or add facts about the main subject's quantity.
\item If you include numbers, they must only represent intermediate or unrelated quantities and must not affect the final calculation.
\item NEVER add a number that equals the final answer
\item Each new sentence must be standalone and insertable
\item Specify exactly WHERE to insert (after which sentence number, counting from 1)
\end{enumerate}

\medskip
\textbf{OUTPUT FORMAT:}

Return a JSON object with a ``sentences'' key containing a list:

\begin{tcolorbox}[
  colback=jsonbg,
  coltext=jsontext,
  colframe=jsonbg,
  fontupper=\ttfamily\small,
  boxrule=0pt,
  arc=1mm,
  left=2mm,
  right=2mm,
  top=2mm,
  bottom=2mm,
]
\{\\
~~"sentences": [\\
~~~~\{"sentence": "Bengali sentence here", "insert\_after": 1\},\\
~~~~\{"sentence": "Another Bengali sentence", "insert\_after": 2\}\\
~~]\\
\}
\end{tcolorbox}

\medskip
\textbf{EXAMPLE 1:}

\textit{Input:}

\textbf{Original:} {\bengalifont রহিমের ৫টি কলম আছে এবং ৩ টি কলম উপহার পেল। সে আরও ৮টি কলম কিনল। রহিমের মোট কত কলম?}

\textbf{Answer:} 16

\textit{Output:}

\begin{tcolorbox}[
  colback=jsonbg,
  coltext=jsontext,
  colframe=jsonbg,
  fontupper=\ttfamily\small,
  boxrule=0pt,
  arc=1mm,
  left=2mm,
  right=2mm,
  top=2mm,
  bottom=2mm,
]
\{\\
~~"sentences": [\\
~~~~\{"sentence": "{\normalfont\bengalifont তার বন্ধু করিমের ১৩টি কলম আছে।}", "insert\_after": 1\},\\
~~~~\{"sentence": "{\normalfont\bengalifont করিমের স্কুলে ৮টি কলম রাখা ছিল।}", "insert\_after": 2\},\\
~~~~\{"sentence": "{\normalfont\bengalifont দোকানে আরও ২০টি কলম ছিল।}", "insert\_after": 3\}\\
~~]\\
\}
\end{tcolorbox}

\medskip
\textbf{EXAMPLE 2:}

\textit{Input:}

\textbf{Original:} {\bengalifont মিনার ৪টি আপেল ছিল। সে বাজারে গেল এবং ৭টি আপেল কিনল। তারপর ৩টি আপেল খেল। মিনার কাছে কতটি আপেল বাকি?}

\textbf{Answer:} 8

\textit{Output:}

\begin{tcolorbox}[
  colback=jsonbg,
  coltext=jsontext,
  colframe=jsonbg,
  fontupper=\ttfamily\small,
  boxrule=0pt,
  arc=1mm,
  left=2mm,
  right=2mm,
  top=2mm,
  bottom=2mm,
]
\{\\
~~"sentences": [\\
~~~~\{"sentence": "{\normalfont\bengalifont বাজারে মোট ১১টি আপেল বিক্রি হয়েছিল সেদিন।}", "insert\_after": 2\},\\
~~~~\{"sentence": "{\normalfont\bengalifont মিনার ভাই রাজুর কাছে ৫৬৭ টি আপেল ছিল।}", "insert\_after": 3\},\\
~~~~\{"sentence": "{\normalfont\bengalifont রাজু বাজার থেকে আরো ১০ টি আপেল কিনলো।}", "insert\_after": 4\}\\
~~]\\
\}
\end{tcolorbox}

\medskip
\textbf{EXAMPLE 3:}

\textit{Input:}

\textbf{Original:} {\bengalifont একটি গাছে ১৫টি পাখি বসেছিল। ৮টি পাখি উড়ে গেল। কতটি পাখি বাকি রইল?}

\textbf{Answer:} 7

\textit{Output:}

\begin{tcolorbox}[
  colback=jsonbg,
  coltext=jsontext,
  colframe=jsonbg,
  fontupper=\ttfamily\small,
  boxrule=0pt,
  arc=1mm,
  left=2mm,
  right=2mm,
  top=2mm,
  bottom=2mm,
]
\{\\
~~"sentences": [\\
~~~~\{"sentence": "{\normalfont\bengalifont পাশের গাছে ২৩টি পাখি ছিল।}", "insert\_after": 1\},\\
~~~~\{"sentence": "{\normalfont\bengalifont আকাশে উড়ন্ত পাখির মধ্যে ৮টি পাখি উত্তর দিকে গেল।}", "insert\_after": 2\}\\
~~]\\
\}
\end{tcolorbox}

\medskip
\textbf{NOW GENERATE:}

\textbf{Original:} \{question\}

\textbf{Answer:} \{answer\}

\textbf{Output (JSON only):}

\end{promptbox}

\subsubsection{Prompt for Generating Orthogonal Attribute Distractor}
\label{app:oad_prompt}
\begin{promptbox}[Orthogonal Attribute Distractor Generation]
\small
\textbf{Your task:} Add 1-3 \textbf{distractor sentences} that give additional properties (price, weight, time, distance, etc.) to objects already in the problem. These properties are in DIFFERENT units than what the question asks and DO NOT change the final answer.

\medskip
\textbf{RULES:}
\begin{enumerate}
\item Add properties like: price ({\bengalifont টাকা}), weight ({\bengalifont কেজি/গ্রাম}), time ({\bengalifont ঘণ্টা/মিনিট}), distance ({\bengalifont কিমি/মিটার}), etc.
\item The question must ask about COUNT or a DIFFERENT property than what you add
\item Properties must be about the SAME objects mentioned in the problem
\item You can use phrases like: ``{\bengalifont যার দাম ক}'', ``{\bengalifont যার ওজন খ}'', ``{\bengalifont যা গ সময় লাগে}''
\end{enumerate}

\medskip
\textbf{OUTPUT FORMAT:}

Return a JSON object with a ``sentences'' key containing a list:

\begin{tcolorbox}[
  colback=jsonbg,
  coltext=jsontext,
  colframe=jsonbg,
  fontupper=\ttfamily\small,
  boxrule=0pt,
  arc=1mm,
  left=2mm,
  right=2mm,
  top=2mm,
  bottom=2mm,
]
\{\\
~~"sentences": [\\
~~~~\{"sentence": "Bengali sentence here", "insert\_after": 1\},\\
~~~~\{"sentence": "Another Bengali sentence", "insert\_after": 2\}\\
~~]\\
\}
\end{tcolorbox}

\medskip
\textbf{EXAMPLE 1:}

\textit{Input:}

\textbf{Original:} {\bengalifont রহিমের ৬টি কলম আছে। সে আরও ৪টি কলম কিনল। রহিমের মোট কত কলম?}

\textbf{Answer:} 10

\textit{Output:}

\begin{tcolorbox}[
  colback=jsonbg,
  coltext=jsontext,
  colframe=jsonbg,
  fontupper=\ttfamily\small,
  boxrule=0pt,
  arc=1mm,
  left=2mm,
  right=2mm,
  top=2mm,
  bottom=2mm,
]
\{\\
~~"sentences": [\\
~~~~\{"sentence": "{\normalfont\bengalifont প্রতিটি কলমের দাম ১৫ টাকা।}", "insert\_after": 1\},\\
~~~~\{"sentence": "{\normalfont\bengalifont ৬ টি কলম দিয়ে ১২০০ পৃষ্ঠা লিখা যায়।}", "insert\_after": 2\}\\
~~]\\
\}
\end{tcolorbox}

\medskip
\textbf{EXAMPLE 2:}

\textit{Input:}

\textbf{Original:} {\bengalifont সেলিমের ৮টি বই ছিল। সে আরও ৫টি বই কিনল। সেলিমের মোট কতটি বই?}

\textbf{Answer:} 13

\textit{Output:}

\begin{tcolorbox}[
  colback=jsonbg,
  coltext=jsontext,
  colframe=jsonbg,
  fontupper=\ttfamily\small,
  boxrule=0pt,
  arc=1mm,
  left=2mm,
  right=2mm,
  top=2mm,
  bottom=2mm,
]
\{\\
~~"sentences": [\\
~~~~\{"sentence": "{\normalfont\bengalifont প্রথম ৮টি বইয়ের মোট ওজন ৩ কেজি।}", "insert\_after": 1\},\\
~~~~\{"sentence": "{\normalfont\bengalifont পরের ৫টি বইয়ের ওজন ১০ কেজি।}", "insert\_after": 2\}\\
~~]\\
\}
\end{tcolorbox}

\medskip
\textbf{EXAMPLE 3:}

\textit{Input:}

\textbf{Original:} {\bengalifont প্রথম দিন রিয়া ১০ কিলোমিটার দৌড়াল। দ্বিতীয় দিন ১৫ কিলোমিটার দৌড়াল। সে মোট কত কিলোমিটার দৌড়ালো?}

\textbf{Answer:} 25

\textit{Output:}

\begin{tcolorbox}[
  colback=jsonbg,
  coltext=jsontext,
  colframe=jsonbg,
  fontupper=\ttfamily\small,
  boxrule=0pt,
  arc=1mm,
  left=2mm,
  right=2mm,
  top=2mm,
  bottom=2mm,
]
\{\\
~~"sentences": [\\
~~~~\{"sentence": "{\normalfont\bengalifont প্রথম দিনের দৌড়ে ২ ঘণ্টা সময় লেগেছিল।}", "insert\_after": 2\},\\
~~~~\{"sentence": "{\normalfont\bengalifont দ্বিতীয় দিনের দৌড়ে ৫ ঘণ্টা সময় লেগেছিল।}", "insert\_after": 3\}\\
~~]\\
\}
\end{tcolorbox}

\medskip
\textbf{NOW GENERATE:}

\textbf{Original:} \{question\}

\textbf{Answer:} \{answer\}

\textbf{Output (JSON only):}

\end{promptbox}

\subsubsection{Prompt for Generating Null-Effect Event Distractor}
\label{app:need_prompt}
\begin{promptbox}[Null-Effect Event Distractor Generation]
\small
\textbf{Your task:} Add 1--3 \textbf{distractor sentences} about actions that were planned but not completed, started but later canceled (net zero effect), or hypothetical situations that are clearly marked as NOT REAL. These sentences MUST NOT change the final answer.

\medskip
\textbf{RULES:}
\begin{enumerate}
\item Actions must involve the MAIN person/entity in the problem and may involve another related person/entity.
\item At least one added sentence must use negation words: ``{\bengalifont কিন্তু না}'', ``{\bengalifont কিন্তু দিলনা}'', ``{\bengalifont রাজি হলো কিন্তু}'', ``{\bengalifont ফিরিয়ে দিল}''\\
   OR hypothetical markers: ``{\bengalifont যদি...হতো}'', ``{\bengalifont তাহলে...হতো}'', ``{\bengalifont কিন্তু ...}''.
\item The actions must have NET ZERO effect on the final quantity (cancelled, returned, reversed, or never actually happened).
\item Make it clear that these actions do not change the final answer.
\end{enumerate}

\medskip
\textbf{OUTPUT FORMAT:}

Return a JSON object with a ``sentences'' key containing a list:

\begin{tcolorbox}[
  colback=jsonbg,
  coltext=jsontext,
  colframe=jsonbg,
  fontupper=\ttfamily\small,
  boxrule=0pt,
  arc=1mm,
  left=2mm,
  right=2mm,
  top=2mm,
  bottom=2mm,
]
\{\\
~~"sentences": [\\
~~~~\{"sentence": "Bengali sentence here", "insert\_after": 1\},\\
~~~~\{"sentence": "Another Bengali sentence", "insert\_after": 2\}\\
~~]\\
\}
\end{tcolorbox}

\medskip
\textbf{EXAMPLE 1:}

\textit{Input:}

\textbf{Original:} {\bengalifont মিনার ১০০টি কলম আছে। প্রতিটি ৫ টাকা দামের। মিনা যদি সব কলম বিক্রি করে, কত টাকা পাবে?}

\textbf{Answer:} 500

\textit{Output:}

\begin{tcolorbox}[
  colback=jsonbg,
  coltext=jsontext,
  colframe=jsonbg,
  fontupper=\ttfamily\small,
  boxrule=0pt,
  arc=1mm,
  left=2mm,
  right=2mm,
  top=2mm,
  bottom=2mm,
]
\{\\
~~"sentences": [\\
~~~~\{"sentence": "{\normalfont\bengalifont মিনার ভাই রাজুর কাছে ২০টি কলম আছে।}", "insert\_after": 1\},\\
~~~~\{"sentence": "{\normalfont\bengalifont মিনা রাজুর কাছে ১৫টি কলম চাইল।}", "insert\_after": 2\},\\
~~~~\{"sentence": "{\normalfont\bengalifont রাজু ১০টি দিতে রাজি হলো, কিন্তু পরে আর দিলনা।}", "insert\_after": 3\}\\
~~]\\
\}
\end{tcolorbox}

\medskip
\textbf{EXAMPLE 2:}

\textit{Input:}

\textbf{Original:} {\bengalifont নাফিসার ২০টি চকলেট ছিল। সে ৮টি চকলেট খেল। কতটি চকলেট বাকি?}

\textbf{Answer:} 12

\textit{Output:}

\begin{tcolorbox}[
  colback=jsonbg,
  coltext=jsontext,
  colframe=jsonbg,
  fontupper=\ttfamily\small,
  boxrule=0pt,
  arc=1mm,
  left=2mm,
  right=2mm,
  top=2mm,
  bottom=2mm,
]
\{\\
~~"sentences": [\\
~~~~\{"sentence": "{\normalfont\bengalifont সে বন্ধুকে ৫টি চকলেট দিয়েছিল।}", "insert\_after": 1\},\\
~~~~\{"sentence": "{\normalfont\bengalifont কিন্তু তার বন্ধু পরে সবকটি চকলেট তাকে ফিরিয়ে দিল।}", "insert\_after": 2\}\\
~~]\\
\}
\end{tcolorbox}

\medskip
\textbf{EXAMPLE 3:}

\textit{Input:}

\textbf{Original:} {\bengalifont করিমের ১২টি আম আছে। সে ৫টি আম কিনল। করিমের মোট কতটি আম?}

\textbf{Answer:} 17

\textit{Output:}

\begin{tcolorbox}[
  colback=jsonbg,
  coltext=jsontext,
  colframe=jsonbg,
  fontupper=\ttfamily\small,
  boxrule=0pt,
  arc=1mm,
  left=2mm,
  right=2mm,
  top=2mm,
  bottom=2mm,
]
\{\\
~~"sentences": [\\
~~~~\{"sentence": "{\normalfont\bengalifont যদি বড় বাজার খোলা থাকত, তাহলে করিম ২০টি আম কিনতে পারত।}", "insert\_after": 1\},\\
~~~~\{"sentence": "{\normalfont\bengalifont কিন্তু বাজার বন্ধ ছিল, তাই সে ২০টি আম কিনতে পারেনি।}", "insert\_after": 2\}\\
~~]\\
\}
\end{tcolorbox}

\medskip
\textbf{NOW GENERATE:}

\textbf{Original:} \{question\}

\textbf{Answer:} \{answer\}

\textbf{Output (JSON only):}

\end{promptbox}

\subsubsection{Prompt for Verifying Modified Problems}
\label{app:gpt_verify}
\begin{verifybox}[MWP Distractor Verification]
\small
You are a math problem verifier. Your task is to check whether the modified Bengali math problem preserves the same final answer as the original.

\medskip
\textbf{INSTRUCTIONS:}
\begin{enumerate}
\item Read the original problem.
\item Read the modified problem.
\item Check whether the modified problem changes the numbers or operations affecting the final answer.
\item Set ``answers\_match'' to true only if the modified problem should have the same numerical answer as the original (and therefore match the ground truth).
\item Provide a brief justification.
\end{enumerate}

\medskip
\textbf{ORIGINAL PROBLEM:}

\{original\}

\medskip
\textbf{MODIFIED PROBLEM:}

\{modified\}

\medskip
\textbf{GROUND TRUTH ANSWER:}

\{ground\_truth\}

\medskip
\textbf{OUTPUT FORMAT (JSON only):}

\begin{tcolorbox}[
  colback=verifyjsonbg,
  coltext=verifyjsontext,
  colframe=verifyjsonbg,
  fontupper=\ttfamily\small,
  boxrule=0pt,
  arc=1mm,
  left=2mm,
  right=2mm,
  top=2mm,
  bottom=2mm,
]
\{\\
~~"justification": "brief explanation",\\
~~"answers\_match": true/false\\
\}
\end{tcolorbox}

\medskip
\textbf{Output (JSON only):}

\end{verifybox}

\subsubsection{Prompt for Generating Computational Graph}
\label{app:graph_prompt}
\begin{graphbox}[Computational Graph Generation]
\small
You are an expert `target language' Math Reasoner. Your task is to solve mathematical problems by constructing a ``Computational Graph''.

\medskip
\textbf{Graph Rules:}
\begin{itemize}
\item \texttt{id}: Unique identifier (e.g., ``n1'', ``n2'').
\item \texttt{val}: The raw number extracted from text (for input nodes).
\item \texttt{op}: The operation (\texttt{add}, \texttt{sub}, \texttt{mul}, \texttt{div}, \texttt{round}, \texttt{sqrt}, \texttt{floor}, \texttt{sum}, \texttt{mean}). Use \texttt{const} for input numbers.
\item \texttt{args}: List of input node IDs.
\item \texttt{distractor}: Boolean (\texttt{true} / \texttt{false}). Set to \texttt{true} if the node is NOT used in the final calculation path.
\item \texttt{label}: Label for the node.
\end{itemize}

\medskip
\textbf{Available Operations:}
\begin{itemize}
\item \textbf{Input:} \texttt{const} (Use this for all numbers found in text or constants).
\item \textbf{Arithmetic:} \texttt{add}, \texttt{sub}, \texttt{mul}, \texttt{div}, \texttt{abs} (absolute difference).
\item \textbf{Logic/Stats:} \texttt{sum}, \texttt{mean}, \texttt{min} (minimum), \texttt{max} (maximum).
\item \textbf{Rounding:} \texttt{round} (nearest int), \texttt{floor} (round down), \texttt{ceil} (round up).
\item \textbf{Advanced:} \texttt{sqrt}, \texttt{pow}, \texttt{mod} (remainder), \texttt{gcd}, \texttt{lcm}.
\item \textbf{Output:} \texttt{identity} (``final\_result'' points to the answer node)
\end{itemize}

\medskip
Only output a JSON graph representing the solution, nothing else. Nodes must be topologically sorted, and there must be exactly one ``final\_result'' node that represents the final answer. One example is provided below.

\medskip
\textbf{Example:}

\textit{Question:}

{\bengalifont মিনার কাছে ১২২১৯৫ টা কলম আছে। রাজুর কাছে ২৫০৮৪ টা কলম আছে। মিনা রাজুর কাছে ১১২৬ টি কলম চাইল। রাজু ১০০০ টি কলম দিতে রাজি হল, কিন্তু পরে আর দিলেনা। প্রতিটি কলমের দাম ৪৫.৬ টাকা। মিনা যদি কলমগুলো বিক্রি করতে চায়, সে কত টাকা পাবে?}

\textit{Output:}

\begin{tcolorbox}[
  colback=graphjsonbg,
  coltext=graphjsontext,
  colframe=graphjsonbg,
  fontupper=\ttfamily\scriptsize,
  boxrule=0pt,
  arc=1mm,
  left=2mm,
  right=2mm,
  top=2mm,
  bottom=2mm,
]
\{\\
~~"nodes": [\\
~~~~\{"id": "n1", "op": "const", "val": 122195, "distractor": false,\\
~~~~~~"label": "{\normalfont\bengalifont মিনার কলম}"\},\\
~~~~\{"id": "n2", "op": "const", "val": 25084, "distractor": true,\\
~~~~~~"label": "{\normalfont\bengalifont রাজুর কলম}"\},\\
~~~~\{"id": "n3", "op": "const", "val": 1126, "distractor": true,\\
~~~~~~"label": "{\normalfont\bengalifont মিনা রাজুর কাছে চাইল}"\},\\
~~~~\{"id": "n4", "op": "const", "val": 1000, "distractor": true,\\
~~~~~~"label": "{\normalfont\bengalifont রাজু দিতে রাজি হল}"\},\\
~~~~\{"id": "n5", "op": "const", "val": 45.6, "distractor": false,\\
~~~~~~"label": "{\normalfont\bengalifont প্রতিটি কলমের দাম}"\},\\
~~~~\{"id": "total\_money", "op": "mul", "args": ["n1", "n5"],\\
~~~~~~"distractor": false, "label": "{\normalfont\bengalifont মিনার মোট টাকা}"\},\\
~~~~\{"id": "final\_result", "op": "identity", "args": ["total\_money"],\\
~~~~~~"distractor": false, "label": "{\normalfont\bengalifont চূড়ান্ত উত্তর}"\}\\
~~]\\
\}
\end{tcolorbox}

\medskip
\textbf{Your Task:}

\textit{Question:}

\{question\}

\textit{Output:}

\end{graphbox}

\newpage
\subsubsection{Prompt for Reasoning Models}
\label{app:prompt_reasoning}
\begin{cotbox}[Reasoning Model Prompt]
\small

\begin{tcolorbox}[
  colback=cotcontentbg,
  colframe=cotcontentborder,
  fontupper=\ttfamily\scriptsize\small,
  boxrule=0.5pt,
  arc=1mm,
  left=2mm,
  right=2mm,
  top=3mm,
  bottom=3mm,
  breakable,
]
Please reason step by step in Bangla, and put your final numeric answer within \textbackslash{}boxed\{\}.

\vspace{3mm}
{\normalfont\bengalifont প্রশ্ন:} \{question\}
\end{tcolorbox}

\end{cotbox}

\subsubsection{Prompt for Chain of Thought (5-Shot)}
\label{app:cot}
\begin{cotbox}[Chain of Thought Prompt]
\small

\begin{tcolorbox}[
  colback=cotcontentbg,
  colframe=cotcontentborder,
  fontupper=\ttfamily\scriptsize\small,
  boxrule=0.5pt,
  arc=1mm,
  left=2mm,
  right=2mm,
  top=3mm,
  bottom=3mm,
  breakable,
]
{\normalfont\bengalifont
প্রশ্ন: রজারের 5টি টেনিস বল আছে। সে আরও 2 ক্যান টেনিস বল কিনেছে। প্রতিটি ক্যানে 3টি করে টেনিস বল আছে। তার কাছে এখন কতগুলি টেনিস বল আছে?

ধাপে ধাপে উত্তর: রজারের প্রথমে 5টি বল ছিল। 2টি ক্যানের প্রতিটিতে 3টে টেনিস বল মানে 6টি টেনিস বল। 5 + 6 = 11। উত্তর হল 11।

\vspace{2mm}
প্রশ্ন: সার্ভার কক্ষে নয়টি কম্পিউটার ছিল। সোমবার থেকে বৃহস্পতিবার প্রতিদিন আরও পাঁচটি করে কম্পিউটার স্থাপন করা হয়েছিল। সার্ভার কক্ষে এখন কতগুলি কম্পিউটার আছে?

ধাপে ধাপে উত্তর: সোমবার থেকে বৃহস্পতিবার 4দিন হয়। প্রতিদিন 5টি করে কম্পিউটার যোগ করা হয়েছে। যার অর্থ মোট 4 * 5 = 20টি কম্পিউটার যোগ করা হয়েছে। শুরুতে 9টি কম্পিউটার ছিল, তাই এখন 9 + 20 = 29টি কম্পিউটার রয়েছে। উত্তর হল 29।

\vspace{2mm}
প্রশ্ন: লিয়ার 32টি চকোলেট ছিল এবং তার বোনের ছিল 42টি। যদি তারা 35টি খেয়ে থাকে, তাহলে তাদের কাছে মোট কতগুলি অবিশিষ্ট আছে?

ধাপে ধাপে উত্তর: লিয়ার 32টি চকোলেট ছিল এবং লিয়ার বোনের ছিল 42টি। যার অর্থ শুরুতে 32 + 42 = 74টি চকোলেট ছিল। 35টি খাওয়া হয়ে গেছে। তাই তাদের কাছে মোট 74 - 35 = 39টি চকোলট আছে। উত্তর হল 39।

\vspace{2mm}
প্রশ্ন: শনের পাঁচটি খেলনা আছে। ক্রিসমাস উপলক্ষে সে তার মাতা ও পিতা উভয়ের থেকে দুটি করে খেলনা পেয়েছে। তার কাছে এখন কতগুলি খেলনা আছে?

ধাপে ধাপে উত্তর: তার কাছে 5টি খেলনা আছে। সে তার মাতার থেকে 2টি খেলনা পেয়েছিল অতএব, এরপর তার 5 + 2 = 7টি খেলনা হয়েছে। তারপর সে তার পিতার থেকে 2টি খেলনা পেয়েছিল, তাই তার মোট 7 + 2 = 9 টি খেলনা হয়েছে। উত্তর হল 9।

\vspace{2mm}
প্রশ্ন: মাইকেলের 58টি গলফ বল ছিল। মঙ্গলবার, সে 23টি গলফ বল হারিয়েছিল। বুধবার, সে আরও 2টি বল হারিয়েছিল। বুধবারের শেষে তার কাছে কয়টি গলফ বল ছিল?

ধাপে ধাপে উত্তর: শুরুতে মাইকেলের কাছে 58টি গলফ বল ছিল এবং সে 23টি বল হারিয়েছিল, তাই তার 58 - 23 = 35টি বল আছে। আরও 2টি বল হারানোর পর, তার এখন 35 - 2 = 33টি বল আছে। উত্তর হল 33।

\vspace{3mm}
প্রশ্ন:} \{question\}\\
{\normalfont\bengalifont ধাপে ধাপে উত্তর:}
\end{tcolorbox}

\end{cotbox}

\begin{table*}[t]
\centering
\subsection{Examples of Augmented Questions from \ourdb}

\label{app:examples}
\scriptsize
\begin{tabular}{p{0.12\textwidth}p{0.38\textwidth}p{0.38\textwidth}}
\toprule
\textbf{Dataset \& Type} & \textbf{Original} & \textbf{Augmented} \\
\midrule

\multicolumn{3}{l}{\textbf{MSVAMP Dataset}} \\
\midrule

RED & 
{\bengalifont জুলিয়া সোমবার ১৮ জন ছেলেমেয়ের সঙ্গে ট্যাগ খেলেন। মঙ্গলবার তিনি ১০ জন ছেলেমেয়ের সঙ্গে ট্যাগ খেলেন। সোমবার তিনি কতজন ছেলেমেয়ের সঙ্গে আরও খেলেন তার সংখ্যা কত?} 

\vspace{1mm}
\textit{(Julia played tag with 18 children on Monday. On Tuesday she played with 10 children. How many more children did she play with on Monday?)}
&
{\bengalifont জুলিয়া সোমবার ১৮ জন ছেলেমেয়ের সঙ্গে ট্যাগ খেলেন। মঙ্গলবার তিনি ১০ জন ছেলেমেয়ের সঙ্গে ট্যাগ খেলেন। \textcolor{red}{মঙ্গলবার মাঠে আরও ১৫ জন ছেলেমেয়ে খেলতে এসেছিল। তার ছোট বোন বুধবার ১২ জন ছেলেমেয়ের সঙ্গে লুকোচুরি খেলেছিল।} সোমবার তিনি কতজন ছেলেমেয়ের সঙ্গে আরও খেলেন তার সংখ্যা কত?}

\vspace{1mm}
\textit{(Julia played tag with 18 children on Monday. On Tuesday she played with 10 children. \textcolor{red}{15 more children came to play on Tuesday. Her younger sister played hide-and-seek with 12 children on Wednesday.} How many more children did she play with on Monday?)}
\\
\midrule

OAD &
{\bengalifont জুলিয়া সোমবার ১৮ জন ছেলেমেয়ের সঙ্গে ট্যাগ খেলেন। মঙ্গলবার তিনি ১০ জন ছেলেমেয়ের সঙ্গে ট্যাগ খেলেন। সোমবার তিনি কতজন ছেলেমেয়ের সঙ্গে আরও খেলেন তার সংখ্যা কত?}

\vspace{1mm}
\textit{(Julia played tag with 18 children on Monday. On Tuesday she played with 10 children. How many more children did she play with on Monday?)}
&
{\bengalifont জুলিয়া সোমবার ১৮ জন ছেলেমেয়ের সঙ্গে ট্যাগ খেলেন। \textcolor{red}{সোমবার খেলতে ১ ঘণ্টা সময় লেগেছিল।} মঙ্গলবার তিনি ১০ জন ছেলেমেয়ের সঙ্গে ট্যাগ খেলেন। \textcolor{red}{মঙ্গলবার খেলতে ৪৫ মিনিট সময় লেগেছিল।} সোমবার তিনি কতজন ছেলেমেয়ের সঙ্গে আরও খেলেন তার সংখ্যা কত?}

\vspace{1mm}
\textit{(Julia played tag with 18 children on Monday. \textcolor{red}{It took 1 hour to play on Monday.} On Tuesday she played with 10 children. \textcolor{red}{It took 45 minutes to play on Tuesday.} How many more children did she play with on Monday?)}
\\
\midrule

NEED &
{\bengalifont জুলিয়া সোমবার ১৮ জন ছেলেমেয়ের সঙ্গে ট্যাগ খেলেন। মঙ্গলবার তিনি ১০ জন ছেলেমেয়ের সঙ্গে ট্যাগ খেলেন। সোমবার তিনি কতজন ছেলেমেয়ের সঙ্গে আরও খেলেন তার সংখ্যা কত?}

\vspace{1mm}
\textit{(Julia played tag with 18 children on Monday. On Tuesday she played with 10 children. How many more children did she play with on Monday?)}
&
{\bengalifont জুলিয়া সোমবার ১৮ জন ছেলেমেয়ের সঙ্গে ট্যাগ খেলেন। \textcolor{red}{জুলিয়া ভেবেছিলেন আরও ৫ জন ছেলেমেয়েকে সোমবার খেলায় ডাকবেন, কিন্তু পরে আর ডাকলেন না।} মঙ্গলবার তিনি ১০ জন ছেলেমেয়ের সঙ্গে ট্যাগ খেলেন। \textcolor{red}{যদি সোমবার বৃষ্টি না হতো, তাহলে আরও ৩ জন যোগ দিত, কিন্তু শেষ পর্যন্ত তারা আসেনি।} সোমবার তিনি কতজন ছেলেমেয়ের সঙ্গে আরও খেলেন তার সংখ্যা কত?}

\vspace{1mm}
\textit{(Julia played tag with 18 children on Monday. \textcolor{red}{Julia thought about inviting 5 more children to the game on Monday, but she didn't call them later.} On Tuesday she played with 10 children. \textcolor{red}{If it hadn't rained on Monday, 3 more would have joined, but they didn't come in the end.} How many more children did she play with on Monday?)}
\\
\midrule

\multicolumn{3}{l}{\textbf{MGSM Dataset}} \\
\midrule

RED &
{\bengalifont সোমবারে স্যাডি 8 ঘণ্টা ঘুমিয়েছিলেন। পরবর্তী দুইদিনের প্রত্যেকদিনে, তিনি 2 ঘণ্টা কম ঘুমিয়েছিলেন কারণ তাঁকে কিছু অ্যাসাইনমেন্ট শেষ করতে হত। যদি তিনি ঐ দুইদিনের তুলনায় বাকি সপ্তাহে 1 ঘণ্টা বেশি ঘুমিয়ে থাকেন, তবে সারা সপ্তাহ তিনি মোট কত ঘণ্টা ঘুমিয়েছিলেন?}

\vspace{1mm}
\textit{(Sadie slept 8 hours on Monday. On each of the next two days, she slept 2 hours less because she had to finish some assignments. If she slept 1 hour more than those two days for the rest of the week, how many hours did she sleep in total for the whole week?)}
&
{\bengalifont সোমবারে স্যাডি 8 ঘণ্টা ঘুমিয়েছিলেন। \textcolor{red}{স্যাডির ছোট বোন মঙ্গলবারে মাত্র ৬ ঘণ্টা ঘুমিয়েছিল।} পরবর্তী দুইদিনের প্রত্যেকদিনে, তিনি 2 ঘণ্টা কম ঘুমিয়েছিলেন কারণ তাঁকে কিছু অ্যাসাইনমেন্ট শেষ করতে হত। \textcolor{red}{বুধবার রাতে স্যাডি ঘুমানোর আগে একটি বই পড়েছিলেন।} যদি তিনি ঐ দুইদিনের তুলনায় বাকি সপ্তাহে 1 ঘণ্টা বেশি ঘুমিয়ে থাকেন, তবে সারা সপ্তাহ তিনি মোট কত ঘণ্টা ঘুমিয়েছিলেন?}

\vspace{1mm}
\textit{(Sadie slept 8 hours on Monday. \textcolor{red}{Sadie's younger sister slept only 6 hours on Tuesday.} On each of the next two days, she slept 2 hours less because she had to finish some assignments. \textcolor{red}{On Wednesday night Sadie read a book before sleeping.} If she slept 1 hour more than those two days for the rest of the week, how many hours did she sleep in total for the whole week?)}
\\
\midrule

OAD &
{\bengalifont সোমবারে স্যাডি 8 ঘণ্টা ঘুমিয়েছিলেন। পরবর্তী দুইদিনের প্রত্যেকদিনে, তিনি 2 ঘণ্টা কম ঘুমিয়েছিলেন কারণ তাঁকে কিছু অ্যাসাইনমেন্ট শেষ করতে হত। যদি তিনি ঐ দুইদিনের তুলনায় বাকি সপ্তাহে 1 ঘণ্টা বেশি ঘুমিয়ে থাকেন, তবে সারা সপ্তাহ তিনি মোট কত ঘণ্টা ঘুমিয়েছিলেন?}

\vspace{1mm}
\textit{(Sadie slept 8 hours on Monday. On each of the next two days, she slept 2 hours less because she had to finish some assignments. If she slept 1 hour more than those two days for the rest of the week, how many hours did she sleep in total for the whole week?)}
&
{\bengalifont সোমবারে স্যাডি 8 ঘণ্টা ঘুমিয়েছিলেন। \textcolor{red}{সোমবারে স্যাডি ঘুমানোর পর ৫০০ ক্যালরি বার্ন করেছিলেন।} পরবর্তী দুইদিনের প্রত্যেকদিনে, তিনি 2 ঘণ্টা কম ঘুমিয়েছিলেন কারণ তাঁকে কিছু অ্যাসাইনমেন্ট শেষ করতে হত। \textcolor{red}{পরবর্তী দুইদিনে প্রতি রাতে ঘুমানোর জন্য তাঁর বিছানার চাদর প্রতিবার ২০ মিনিটে পরিষ্কার করা হয়।} যদি তিনি ঐ দুইদিনের তুলনায় বাকি সপ্তাহে 1 ঘণ্টা বেশি ঘুমিয়ে থাকেন, তবে সারা সপ্তাহ তিনি মোট কত ঘণ্টা ঘুমিয়েছিলেন?}

\vspace{1mm}
\textit{(Sadie slept 8 hours on Monday. \textcolor{red}{Sadie burned 500 calories after sleeping on Monday.} On each of the next two days, she slept 2 hours less because she had to finish some assignments. \textcolor{red}{On the next two days, her bed sheet is cleaned in 20 minutes each night for sleeping.} If she slept 1 hour more than those two days for the rest of the week, how many hours did she sleep in total for the whole week?)}
\\
\midrule

NEED &
{\bengalifont সোমবারে স্যাডি 8 ঘণ্টা ঘুমিয়েছিলেন। পরবর্তী দুইদিনের প্রত্যেকদিনে, তিনি 2 ঘণ্টা কম ঘুমিয়েছিলেন কারণ তাঁকে কিছু অ্যাসাইনমেন্ট শেষ করতে হত। যদি তিনি ঐ দুইদিনের তুলনায় বাকি সপ্তাহে 1 ঘণ্টা বেশি ঘুমিয়ে থাকেন, তবে সারা সপ্তাহ তিনি মোট কত ঘণ্টা ঘুমিয়েছিলেন?}

\vspace{1mm}
\textit{(Sadie slept 8 hours on Monday. On each of the next two days, she slept 2 hours less because she had to finish some assignments. If she slept 1 hour more than those two days for the rest of the week, how many hours did she sleep in total for the whole week?)}
&
{\bengalifont সোমবারে স্যাডি 8 ঘণ্টা ঘুমিয়েছিলেন। \textcolor{red}{স্যাডি প্রথমে ঠিক করেছিলেন যে মঙ্গলবার ৮ ঘণ্টা ঘুমাবেন, কিন্তু না, পরে সিদ্ধান্ত বদলান এবং ৬ ঘণ্টা ঘুমান।} পরবর্তী দুইদিনের প্রত্যেকদিনে, তিনি 2 ঘণ্টা কম ঘুমিয়েছিলেন কারণ তাঁকে কিছু অ্যাসাইনমেন্ট শেষ করতে হত। \textcolor{red}{বুধবার তিনি ভাবছিলেন আরও কম ঘুমাবেন, কিন্তু শেষ পর্যন্ত আগের দিনের মতোই ৬ ঘণ্টা ঘুমান।} যদি তিনি ঐ দুইদিনের তুলনায় বাকি সপ্তাহে 1 ঘণ্টা বেশি ঘুমিয়ে থাকেন, তবে সারা সপ্তাহ তিনি মোট কত ঘণ্টা ঘুমিয়েছিলেন?}

\vspace{1mm}
\textit{(Sadie slept 8 hours on Monday. \textcolor{red}{Sadie first decided to sleep 8 hours on Tuesday, but no, she changed her mind and slept 6 hours.} On each of the next two days, she slept 2 hours less because she had to finish some assignments. \textcolor{red}{On Wednesday she thought she would sleep even less, but eventually slept 6 hours like the previous day.} If she slept 1 hour more than those two days for the rest of the week, how many hours did she sleep in total for the whole week?)}
\\

\bottomrule
\end{tabular}
\caption{Examples of original and augmented questions with different distractor types across MSVAMP and MGSM datasets. Distractor sentences are marked in \textcolor{red}{red}.}
\label{tab:augmented_examples}
\end{table*}

\end{document}